\documentclass[letterpaper, 10 pt, conference]{ieeeconf}
\IEEEoverridecommandlockouts 
\let\labelindent\relax
\usepackage{hyperref, url, amsmath, amssymb, booktabs, graphicx, bm, color, algorithm, tabularx, enumitem, balance}
\usepackage[noend]{algpseudocode}
\graphicspath{{images/}}

\usepackage{microtype}
\renewcommand{\baselinestretch}{0.98}

\newcolumntype{Y}{>{\centering\arraybackslash}X}

\title{\LARGE \bf Assistive Gym: A Physics Simulation Framework for Assistive Robotics}

\author{Zackory Erickson, Vamsee Gangaram, Ariel Kapusta, C. Karen Liu, and Charles C. Kemp
\thanks{Zackory Erickson, Vamsee Gangaram, and Charles C. Kemp are with the Healthcare Robotics Lab, Georgia Institute of Technology, Atlanta, GA., USA.}%
\thanks{C. Karen Liu is with the Department of Computer Science, Stanford University, Stanford, CA., USA.}%
\thanks{Ariel Kapusta is with Enway GmbH, Berlin, Germany.}%
\thanks{Zackory Erickson is the corresponding author {\tt\footnotesize zackory@gatech.edu}.}%
}

\begin{document}

\maketitle
\thispagestyle{empty}
\pagestyle{empty}

\begin{abstract}
Autonomous robots have the potential to serve as versatile caregivers that improve quality of life for millions of people worldwide. Yet, conducting research in this area presents numerous challenges, including the risks of physical interaction between people and robots. Physics simulations have been used to optimize and train robots for physical assistance, but have typically focused on a single task. In this paper, we present Assistive Gym, an open source physics simulation framework for assistive robots that models multiple tasks. It includes six simulated environments in which a robotic manipulator can attempt to assist a person with activities of daily living (ADLs): itch scratching, drinking, feeding, body manipulation, dressing, and bathing. Assistive Gym models a person's physical capabilities and preferences for assistance, which are used to provide a reward function. We present baseline policies trained using reinforcement learning for four different commercial robots in the six environments. We demonstrate that modeling human motion results in better assistance and we compare the performance of different robots. Overall, we show that Assistive Gym is a promising tool for assistive robotics research. 
\end{abstract}

\section{Introduction}
\label{sec:intro}



In 2014, 27.2 percent, or 85 million, of people living in the United States had a disability~\cite{taylor2014americans}. About 17.6 percent, or 55 million people, had a severe disability. Autonomous robots that provide versatile physical assistance offer an opportunity to positively impact the lives of people who require support with everyday tasks, yet conducting this type of research presents several challenges, including high costs and risks associated with physical human-robot interaction.

When compared to real-world robotic systems, physics simulation allows robots to safely make and learn from mistakes without putting real people at risk. Physics simulations can also parallelize data collection to perform thousands of human-robot trials in a few hours and provide models of people representing a wide spectrum of human body shapes, weights, and physical capabilities/impairments.

\begin{figure}
\centering
\includegraphics[width=0.23\textwidth, trim={10cm 8cm 11.5cm 3cm}, clip]{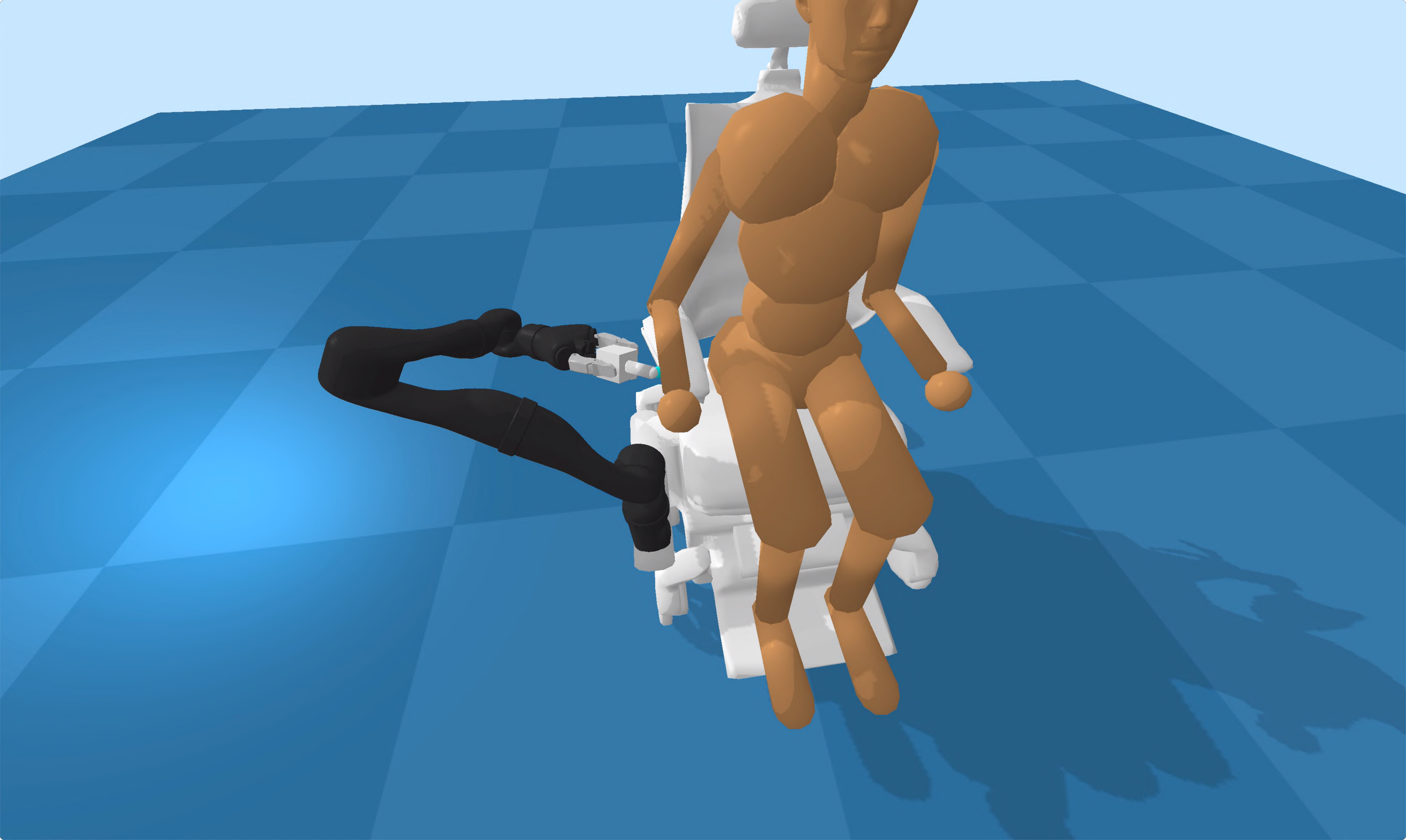}
\includegraphics[width=0.23\textwidth, trim={7cm 8cm 10cm 0cm}, clip]{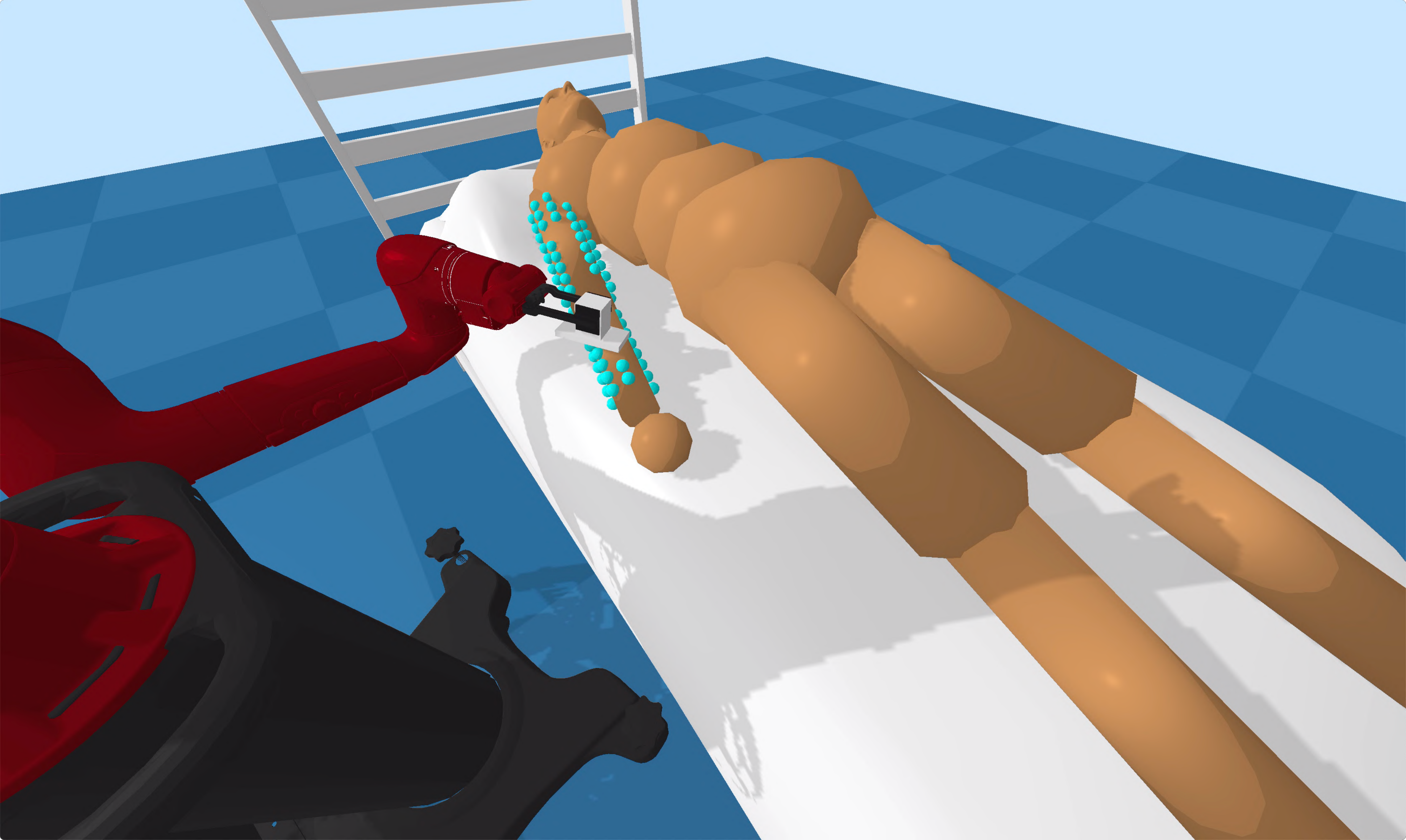}\\[0.1cm]
\ \includegraphics[width=0.23\textwidth, trim={6.5cm 5cm 13.5cm 5cm}, clip]{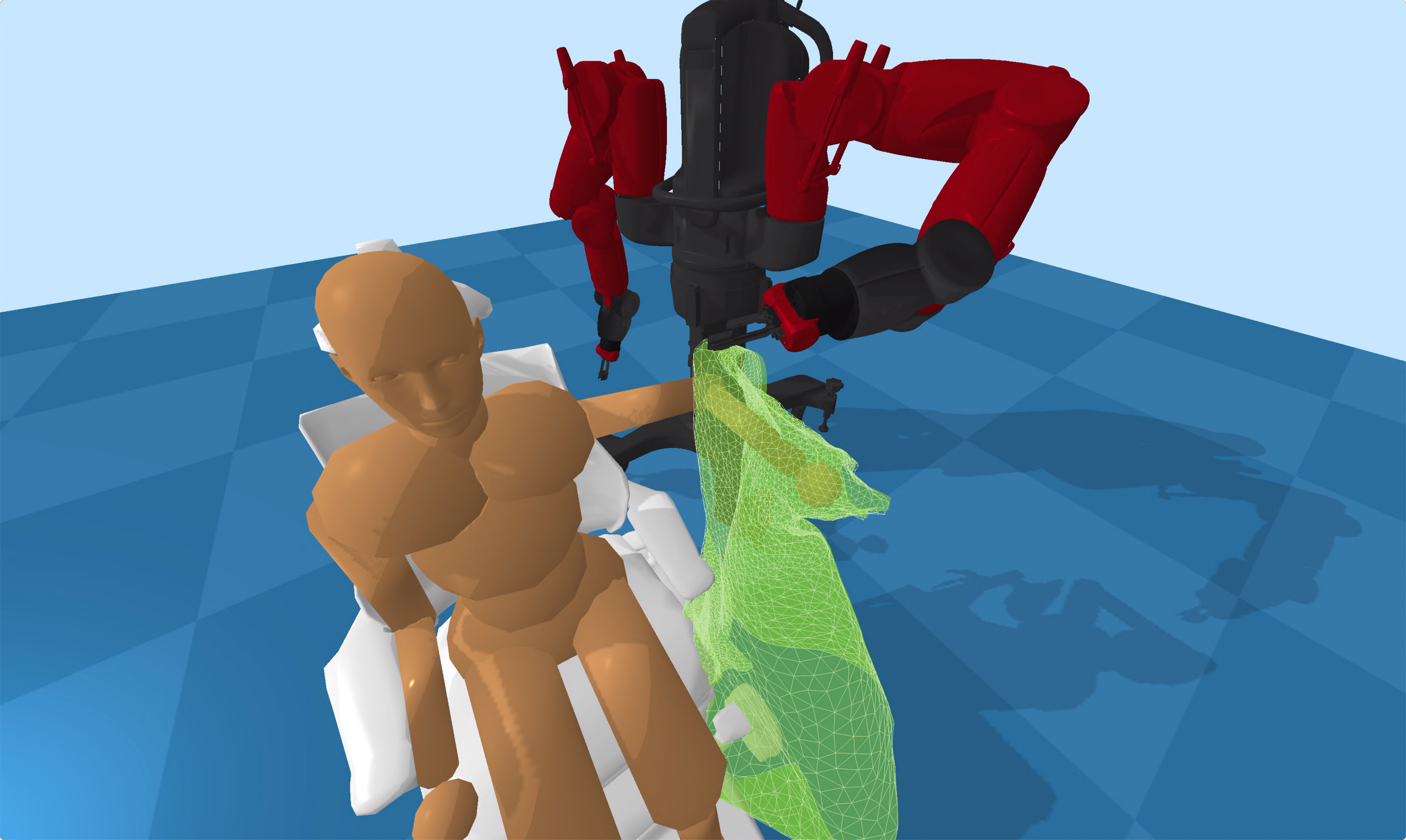}
\includegraphics[width=0.23\textwidth, trim={16cm 12cm 15cm 5cm}, clip]{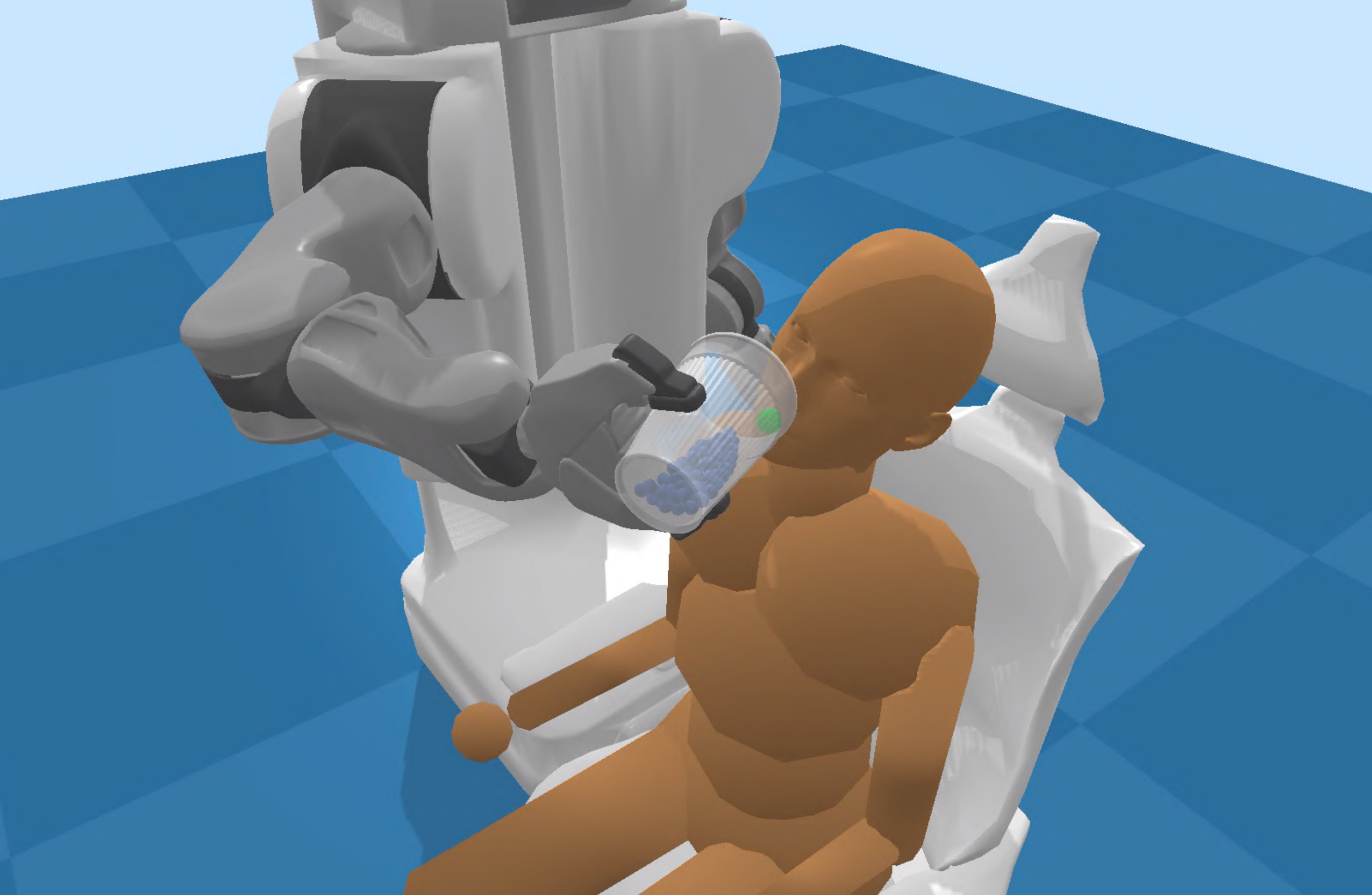}
\vspace{-0.2cm}
\caption{\label{fig:intro}Four collaborative robots in Assistive Gym providing physical assistance. The four tasks include itch scratching, bed bathing, dressing, and drinking assistance.}
\vspace{-0.4cm}
\end{figure}

In this paper, we present Assistive Gym\footnote{\scriptsize\url{https://github.com/Healthcare-Robotics/assistive-gym}}, an open source physics-based simulation framework for physical human-robot interaction and robotic assistance.
In comparison to existing robotic simulation environments, Assistive Gym places a strong emphasis on modeling the interaction between robots and humans (Fig.~\ref{fig:intro}), and builds off of prior research on how robots can provide intelligent physical assistance to people~\cite{chen2013robotshumanity, park2017multimodal, erickson2019multidimensional, schroer2015drinking, kapusta2019dressing}.

Assistive Gym is integrated into the OpenAI Gym framework. It provides an intuitive and familiar interface for developing control algorithms that enable robots to more intelligently interact with and assist people~\cite{brockman2016openai}.
With this framework, we introduce physics-based environments where robots can assist people with six different tasks associated with activities of daily living (ADLs); including itch scratching, drinking, feeding, body manipulation, dressing, and bed bathing assistance---all of which are commonly requested tasks among adults who require physical assistance~\cite{mitzner2014identifying}. 

Assistive Gym presents several use cases for the research community. First, Assistive Gym can be used as a benchmark to compare control algorithms for robots that interact with people. 
Assistive Gym also provides the groundwork for researchers to develop environments and controllers for their own assistive tasks. Additionally, it allows researchers to design or compare robots for a given task.

Understanding a person's preferences for receiving care can be crucial for robots that wish to provide consistent and high quality care. 
We emphasize and model human preferences throughout the Assistive Gym framework, and we demonstrate how robots can learn to prioritize providing assistance that is consistent with a person's preferences.

We provide and evaluate baseline control policies for four commercial robots for each of the assistive tasks: the PR2, Sawyer, Baxter, and Jaco robots. 
Finally, we note that there are often times when a person who requires assistance will have some form of limited motor functionality. 
In these scenarios, we demonstrate that by modeling human movement in simulation, robots learned to provide better assistance, with task success increasing by 30.4\% on average.
We model this as a co-optimization problem, in which both the human and robot are active agents that are simultaneously trained to achieve the same assistive task.

Through this work, we make the following contributions:
\begin{itemize}
\item We release a simulation framework, Assistive Gym, for physical human-robot interaction and robotic assistance.
\item We introduce assistive robotic environments grounded in prior research for six activities of daily living.
\item We present and analyze baseline robot controllers for each assistive task, and we show how Assistive Gym may be used to compare robots for assistive tasks.
\item We show that by modeling human motion through co-optimization, robots can learn to provide improved assistance for a variety of tasks.
\item We discuss modeling of human preferences in physical human-robot interaction simulations and we present a unified model of human preferences for assistive tasks.
\end{itemize}

\section{Related Work}
\label{sec:related_work}

\subsection{Simulation Environments}
OpenAI Gym is a framework for learning control policies for simulated agents that includes a collection of benchmark problems, a common interface, and comparison tools. Its benchmark environments include Atari games and physics-based locomotion agents~\cite{brockman2016openai, duan2016benchmarking, plappert2018multi}. Assistive Gym builds on the OpenAI Gym framework and is intended to fulfill a similar role for the field of assistive robotics research.

Three physics engines commonly used for simulating robotics environments in OpenAI Gym are PyBullet~\cite{coumanspybullet}, DART~\cite{lee2018dart}, and MuJoCo~\cite{todorov2012mujoco}. PyBullet, which we use to build our simulation environments, is a Python module for the open source Bullet Physics Engine that has been used for training and validating real robots using physics simulations~\cite{tan2018sim, zeng2019tossingbot, sadeghi2017sim2real, bousmalis2018using}.

Several recent projects have begun presenting simulation environments for various robotic tasks including manipulation, navigation, or visual tasks~\cite{fan2018surreal, savva2019habitat, kolve2017ai2}. 
Zamora et al. extended OpenAI Gym to ROS and Gazebo~\cite{zamora2016extending}, which has been used by~\cite{rodriguez2019deep}.
Similarly, Fan et al. introduced SURREAL, a physics simulation framework for robotic manipulation research that included six manipulation tasks, such as block stacking and bin picking~\cite{fan2018surreal}. In contrast to existing robotic simulation environments, Assistive Gym provides human-centric environments in which robots learn to directly help people in a variety of tasks that have been shown to be important for quality of life. 

\subsection{Robotic Assistance}

A number of works have explored robot-assisted feeding, using a variety of robots from wheelchair-mounted arms to mobile manipulators~\cite{gallenberger2019transfer, canal2016personalization, rhodes2018robot, naotunna2015meal, park2018multimodal, park2017multimodal}. Various robotic arms have also been used for robotic drinking assistance~\cite{schroer2015drinking, goldau2019autonomous}.
Itch scratching assistance can be especially valuable for people who are unable to move their upper bodies due to disabilities~\cite{chen2013robotshumanity, grice2012wouse}. 
Researchers have also investigated robot assistance for bed bathing~\cite{king2010towards}, including capacitive sensing to sense the human body~\cite{erickson2019multidimensional}.

Robot-assisted dressing has received significant focus in recent years~\cite{klee2015personalized, yamazaki2014bottom, chance2018elbows, pignat2017learning, canal2018joining, tamei2011reinforcement, koganti2016bayesian, gao2016iterative, zhang2019probabilistic}.
Our prior research has also explored the use of physics simulations for learning control strategies for robot-assisted dressing~\cite{erickson2017does, kapusta2019dressing, yu2017haptic, erickson2018deep}. 
Clegg et al. presented a co-optimization approach for a KUKA IIWA robot and person to jointly learn an assistive dressing task in simulation~\cite{cleggdissertation}. We incorporate this co-optimization technique into our evaluations of Assistive Gym (Section~\ref{sec:collaborative}), and we show how this approach can be extended to multiple different robots providing assistance to people for a variety of tasks.

Assistive Gym builds on many of the assistive robotics research discussed above, providing environments for robots to safely learn to interact with people with disabilities at scales much larger than currently available with robots in the real-world.
Assistive Gym aims to facilitate assistive robotics research by providing a common simulation framework that can be used for training algorithms in many common assistive tasks. The framework can lower the barrier to development of algorithms, provide baselines for comparison, and provide canonical problems for the field.

\section{Assistive Gym}
\label{sec:assistive_gym}

Assistive Gym is a simulation framework with high level interfaces for building and customizing simulation environments for robots that physically interact with and assist people.
Assistive Gym environments are built in the open source PyBullet physics engine~\cite{coumanspybullet}. PyBullet presents several benefits for simulating physical human-robot interaction, including real time simulation on both CPUs and GPUs, soft bodies and cloth simulation, and the ability to programmatically create robots and human models of varying shapes, sizes, weights, and joint limits. Assistive Gym integrates directly into the OpenAI Gym interface, allowing for the use of existing control policy learning algorithms, such as deep reinforcement learning.

\subsection{Human Model and Robots}

Assistive Gym provides support for four commercial collaborative robots that are commonly used for physical human-robot interaction. These include the PR2, Jaco, Baxter, and Sawyer robots, as shown in Fig.~\ref{fig:intro}.

We provide default male and female human models, with body sizes, weights, and joint limits matching published 50th percentile values~\cite{tilley2002measure}.
The human model is programmatically generated, allowing for easy modification to the shape and properties of the human.
Self collision between the various limbs and body parts of the human is also enabled. In total, the human model has 40 controllable joints, including an actuated head, torso, arms, waist, and legs. 

Assistive Gym also provides models of several human limitations, including head and arm tremor, joint weakness, and limited range of motion. Given a target joint configuration, $\bm{\bar{q}}$, for the person's arm or head, we model tremor by adding an oscillating offset such that $\bm{q}_t = \bm{\bar{q}} + \epsilon (-1)^{t \bmod 2}$ where $\epsilon \in U(0, 20^{\circ})$ is sampled upon creation of the human model. We model joint weakness by multiplying the maximum torque, $\tau_{max}$, of each human joint by a scaling factor $\beta \in U(0.25, 1)$. We model limited range of motion by multiplying the pose-independent limits for each human joint, $\bm{l}_{min}$, which are negative, and $\bm{l}_{max}$, which are positive, by a scaling factor ${\gamma \in U(0.5, 1)}$.
For the six assistive environments, the simulator randomly selects either the male or female human model and randomly applies one of these three limitations to the human upon creation.

\subsection{Assistive Tasks (Environments)}
\label{sec:envs}

As a part of Assistive Gym, we are releasing a suite of simulation environments for six tasks associated with activities of daily living (ADLs)~\cite{lawton1969assessment}, including:

\begin{itemize}[leftmargin=*]
\item \textbf{Itch Scratching}: A robot holds a small scratching tool and must reach towards a random target scratching location along a person's right arm. 
The robot is rewarded for moving its end effector close to the target location, applying less than 10~N of force near the target, and performing small scratching motions near the target.
\item \textbf{Bed Bathing}: A person lies on a bed in a random resting position while a robot must use a washcloth tool to clean off a person's right arm. 
The robot is rewarded for moving its end effector closer to the person's body and for wiping the bottom of the washcloth tool along the surface of the person's arm. We uniformly distribute markers (3~cm apart) around the person's right arm, for the robot to wipe off.
\item \textbf{Drinking Water}: A robot holds a cup of small spheres (particles) representing water and must help a person drink this water. We randomize the starting orientation of the person's head.
The robot is rewarded for moving the cup closer to the person's mouth, for tilting the cup, and for pouring water into the person's mouth.
\item \textbf{Feeding}: A robot holds a spoon with small spheres representing food on the spoon and must bring this food to a person's mouth without spilling it. We randomize the starting orientation of the person's head.
The robot is rewarded for moving the spoon closer to the person's mouth and placing food into the person's mouth.
\item \textbf{Dressing}: A robot holds a hospital gown and must pull the sleeve of the gown up a person's left forearm and upper arm. 
The robot is rewarded for pulling the opening of the sleeve toward a person's hand, for pulling the sleeve onto a person's arm, and for pulling the sleeve along the central axis of the arm, towards the person's shoulder.
\item \textbf{Arm Manipulation}: A person lies on a bed in a random resting position, with his/her right arm hanging off of the bed. A robot uses a scooping tool to lift a person's arm and place it on the bed next to his/her torso. 
The robot is rewarded for moving its end effectors towards the person's arm and for lifting the person's arm towards their torso.
\end{itemize}

For most of the assistive tasks, a robot holds a task relevant tool, such as a washcloth tool (see ~\cite{erickson2019multidimensional}) or a cup of water. Prior to the start of each task, the robot's base pose with respect to the person is optimized (section~\ref{sec:toc}) and a random perturbation is added to the starting position of the robot's end effector (up to 5~cm along each global axis).

\subsection{Realistic Human Joint Limits}
\label{sec:joint_limits}

\begin{figure}
\centering
\includegraphics[width=0.23\textwidth, trim={15.5cm 15cm 15.5cm 5.4cm}, clip]{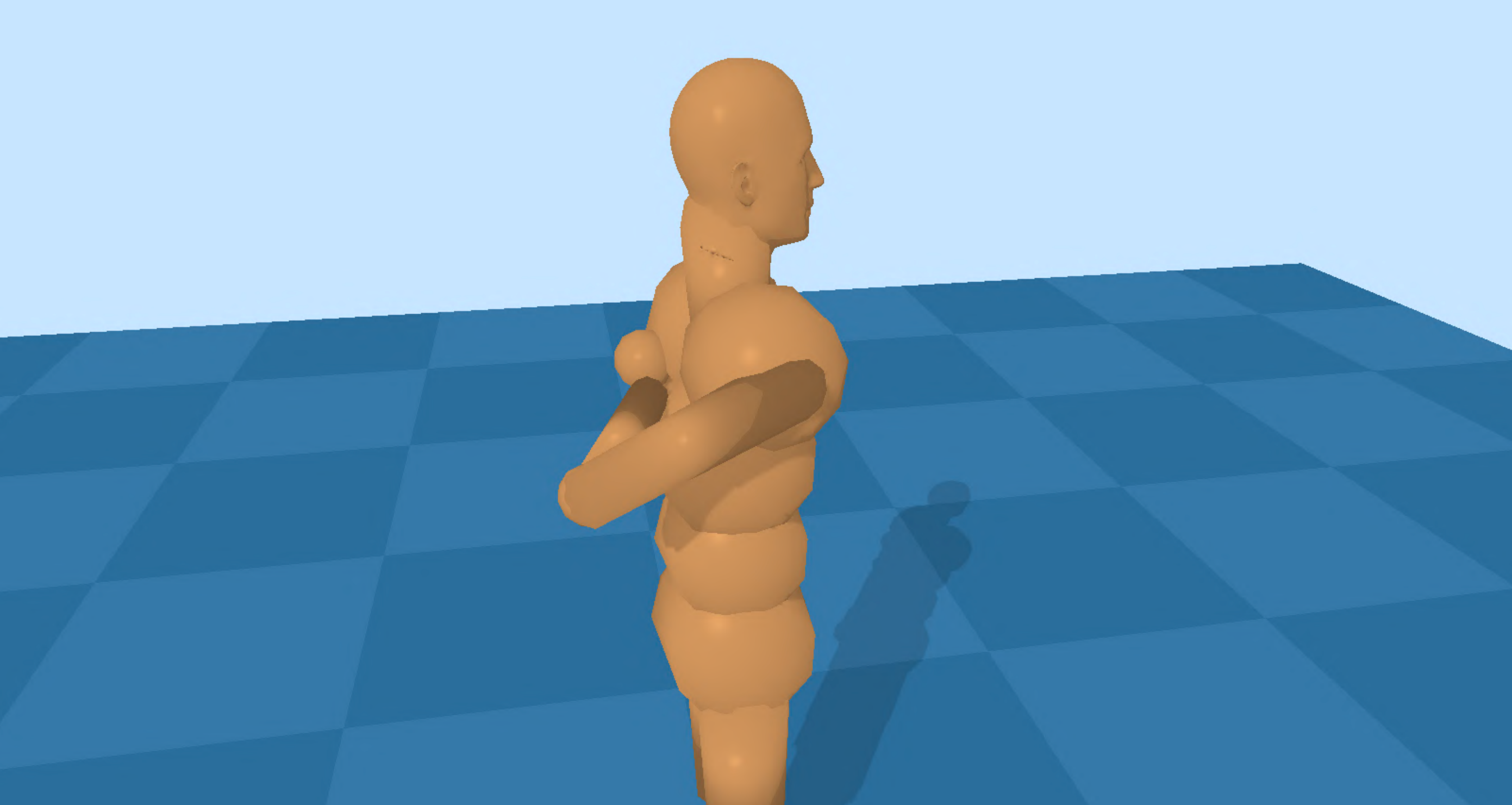}
\includegraphics[width=0.23\textwidth, trim={14cm 12.75cm 12cm 5cm}, clip]{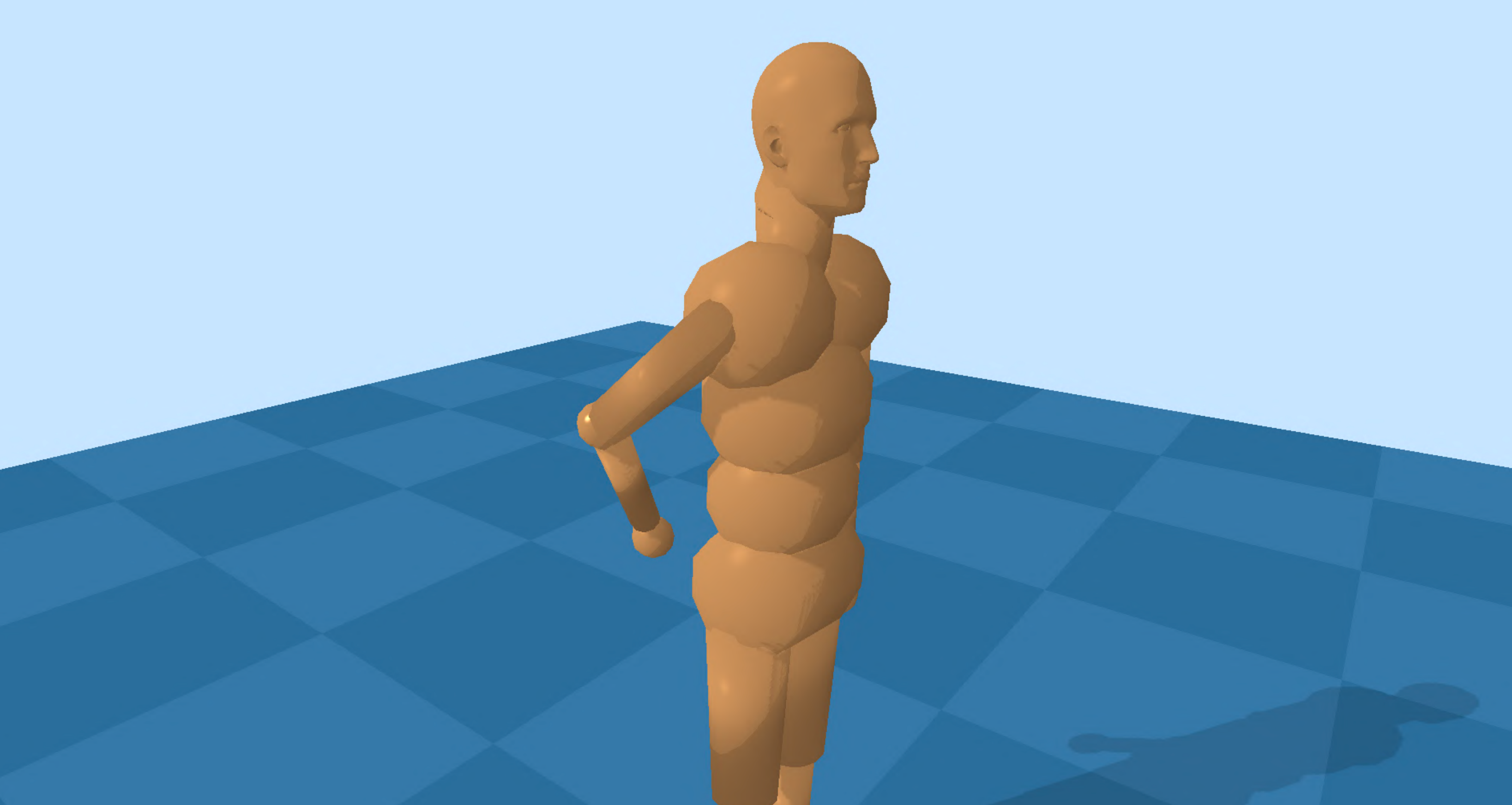}
\vspace{-0.2cm}
\caption{\label{fig:joint_limits}(Left) An attainable arm joint pose when joint limits are pose-independent. (Right) The realistic joint limits reached when the human attempts to move to the same pose with pose-dependent limits enabled.}
\vspace{-0.6cm}
\end{figure}

With physical human-robot interaction in simulation, it is important to model realistic human joint limits as we want robots to learn to provide safe assistance that does not create discomfort. However, modeling joint limits can be difficult since they are pose-dependent---the range of motion for one joint is dependent on the configurations of other joints.

Akhter and Black presented a procedure, consisting of discrete operations, to model realistic human joint limits~\cite{akhter2015pose}. Given a joint configuration for the human arm, $\bm{q}$, the authors provide a function, $C(\bm{q})$, which estimates a binary value for whether the arm configuration is in a valid pose. The authors fit their procedure on human motion capture data.
Recently, Jiang et al. presented an approach to reduce the computational requirements of determining whether a human pose is valid using a fully connected neural network model trained on the human motion capture data~\cite{jiang2018data}. We have incorporated this neural network model into Assistive Gym to model realistic human arm joint limits.

Specifically, at each time step, $t$, during simulation, we compute $C(\bm{q}^{}_{right, t})$ and $C(\bm{q}^{}_{left, t})$ for both the right and left human arm, respectively. If the output from the model indicates that an arm is in an invalid pose, we then set the arm to the joint configuration from the previous time step, $\bm{q}^{}_{\bm{\cdot}, t-1}$, in which the arm was in a valid joint configuration. Additional details on the model and training procedure for realistic joint limits can be found in~\cite{jiang2018data}.

In Fig.~\ref{fig:joint_limits}, we depict the impact of realistic human joint limits in Assistive Gym.
Fig.~\ref{fig:joint_limits} (Left) shows an achievable arm pose that may be uncomfortable when all joints and joint limits are considered independent. Fig.~\ref{fig:joint_limits} (Right) shows an arm pose that is more likely to be comfortable, which is achieved when the realistic joint limit model is enabled and the human attempts to reach the previous configuration.

\subsection{Robot Base Positioning}
\label{sec:toc}
With a well-chosen base pose, a robot can better perform an assistive task despite model error, pose uncertainty, and other sources of variation. We created a baseline method for selecting robot base poses with respect to a person (2D position and orientation) based on task-centric optimization of robot configurations (TOC), particularly joint-limit-weighted kinematic isotropy (JLWKI) and concepts from task-centric manipulability as described in \cite{kapusta2019task}. 

Assistive Gym randomly samples 100 base poses for the robot and attempts to find an inverse-kinematics (IK) solution to each goal end effector pose. It selects the robot base pose with a collision-free IK solution to the most goals. In case of ties, Assistive Gym selects the base pose with highest JLWKI summed across all goal end effector poses. 

JLWKI, as presented by Kapusta and Kemp \cite{kapusta2019task}, is a modification of manipulability from \cite{klein1987dexterity} and kinematic isotropy from \cite{kim1991dexterity}. 
Details on JLWKI are present in \cite{kapusta2019task}.

The base pose for the PR2, Baxter, and Sawyer are optimized at the beginning of a simulation trial for all six assistive tasks in Assistive Gym. For the four tasks with a wheelchair, namely itch scratching, drinking, feeding, and dressing, we mount the Jaco to a fixed location on the wheelchair comparable to the wheelchair location recommended by Kinova who produces the Jaco arm~\cite{kinova}. For the bed bathing and arm manipulation tasks, where the person is lying on a bed, the Jaco robot is mounted to a nightstand next to the bed, and the nightstand pose is optimized.

\subsection{Modeling Human Preferences}
\label{sec:prefs}

\begin{figure}
\centering
\includegraphics[width=0.48\textwidth, trim={0cm 0cm 0cm 0cm}, clip]{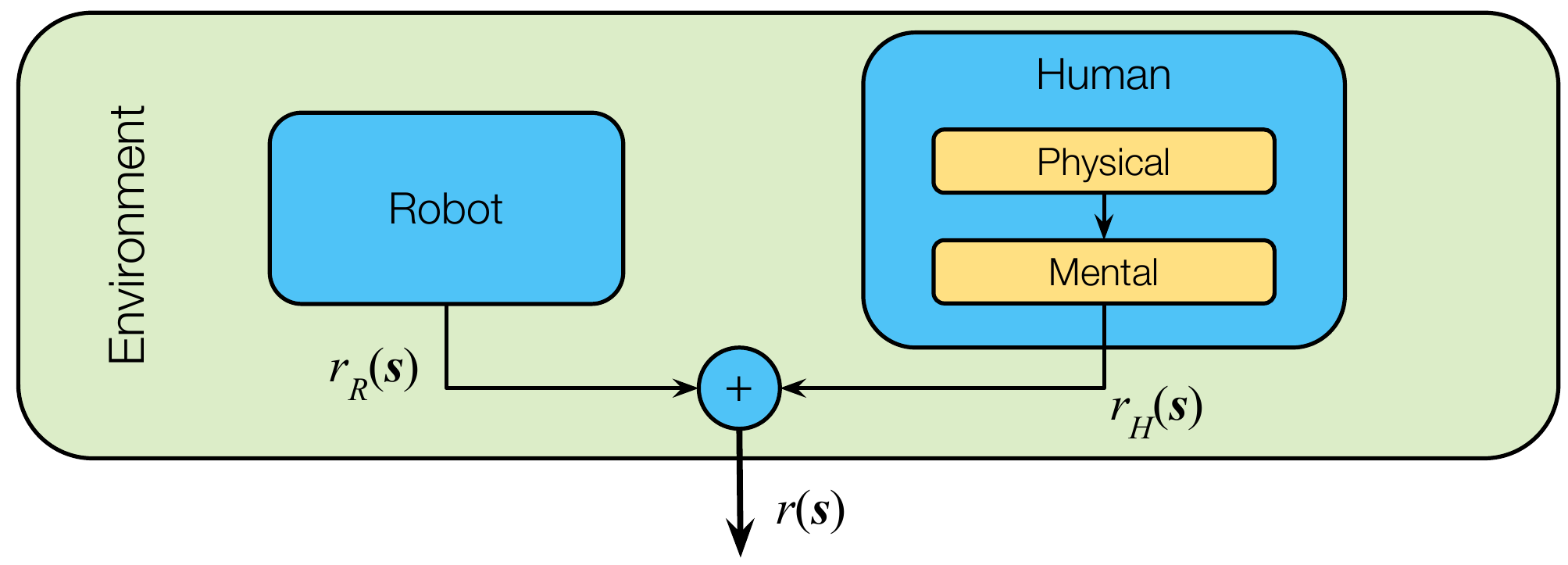}
\vspace{-0.6cm}
\caption{\label{fig:human_prefs} A model of a human-robot interaction environment that includes human preference. The physical state of a person (e.g. amount of force applied to the person) affects the person's mental model. Given an approximate mental state of the person, we can compute a human preference reward $r^{}_H(\bm{s})$, which we combine with the robot's task success reward, $r^{}_R(\bm{s})$.}
\vspace{-0.6cm}
\end{figure}

Understanding a person's preferences for receiving care may improve quality and consistency of care. For example, a person may prefer for the robot not to apply large forces to their body, or for the robot to perform slow and interpretable actions~\cite{dragan2013legibility}. Prior research has noted the importance of considering human preference when planning robot actions such as navigation \cite{kruse2013human, sisbot2010synthesizing} and manipulation \cite{mainprice2011planning, cakmak2011human, dragan2013legibility} around humans. 
Other prior research on reinforcement learning has used human preference feedback to allow learning when reward functions are poorly defined \cite{christiano2017deep} or to increase learning efficiency \cite{pinsler2018sample}. In comparison, we aim to model a set of human preferences that function across a variety of assistive tasks.
We provide a set of baseline human preferences in Assistive Gym which are unified across tasks.

At each time step, Assistive Gym computes a human preference reward, $r^{}_H(\bm{s})$, based on how well a robot is satisfying the person's preferences given the state of the system, $\bm{s}$.
As depicted in Fig.~\ref{fig:human_prefs}, we combine this human preference reward with the robot's task success reward, $r^{}_R(\bm{s})$, to obtain an overall reward, $r(\bm{s})$. By maximizing this reward a robot is able to learn control policies that are consistent with a person's preferences for receiving care. We use this same reward function, $r(\bm{s})$, when training both robot and human policies (Section~\ref{sec:collaborative}). We define $r^{}_H(\bm{s})$ as,
\begin{multline*}
r^{}_H(\bm{s}) = -\bm{\alpha}\bm{\cdot}\bm{\omega}\odot [C^{}_{v}(\bm{s}), C^{}_{f}(\bm{s}), C^{}_{hf}(\bm{s}), \\
C^{}_{fd}(\bm{s}), C^{}_{fdv}(\bm{s}), C^{}_{d}(\bm{s}), C^{}_{p}(\bm{s})].
\end{multline*}

$\bm{\alpha}$ is a vector of activations in $\{0, 1\}$ depicting which human preferences are enabled for a given task, whereas $\bm{\omega}$ represents a vector of weights for each preference.
For our environments, we use $\bm{\omega} = [0.25, 0.01, 0.05, 1.0, 1.0, 0.01, 0.01]$.
$C^{}_{\bm{\cdot}}(\bm{s})$ represents the cost of deviating from human preference in the state of the system, $\bm{s}$. 
We define the penalty terms as,
\begin{itemize}
    \item $C^{}_{v}(\bm{s})$, cost for high end effector velocities.
    \item $C^{}_{f}(\bm{s})$, applying force away from the target assistance location (e.g. human mouth for drinking assistance).
    \item $C^{}_{hf}(\bm{s})$, applying high forces near the target ($>$ 10~N).
    \item $C^{}_{fd}(\bm{s})$, spilling food or water on the person.
    \item $C^{}_{fdv}(\bm{s})$, food / water entering mouth at high velocities.
    \item $C^{}_{d}(\bm{s})$, fabric garments applying force to the body.
    \item $C^{}_{p}(\bm{s})$, applying high pressure with large tools. 
\end{itemize}
For example, $C^{}_{v}(\bm{s}) = \left\Vert \bm{v}^{}_L \right\Vert^{}_2 + \left\Vert \bm{v}^{}_R \right\Vert^{}_2$, where $\bm{v}^{}_L$ and $\bm{v}^{}_R$ are the velocities of the robot's left and right end effectors, respectively, or $\bm{v}^{}_L = \bm{v}^{}_R$ for single arm robots. We include a separate term for high pressure, as a robot must on occasion apply large forces distributed over a large surface area for tasks such as arm manipulation. Full equations for each term can be found in the source code. 

\begin{figure*}
\centering
\raisebox{0.25in}{\rotatebox[origin=t]{90}{\parbox[][0.75cm][t]{1.75cm}{\centering Scratch Itch \\ Jaco}}}
\includegraphics[width=0.23\textwidth, trim={15cm 13.5cm 15cm 9cm}, clip]{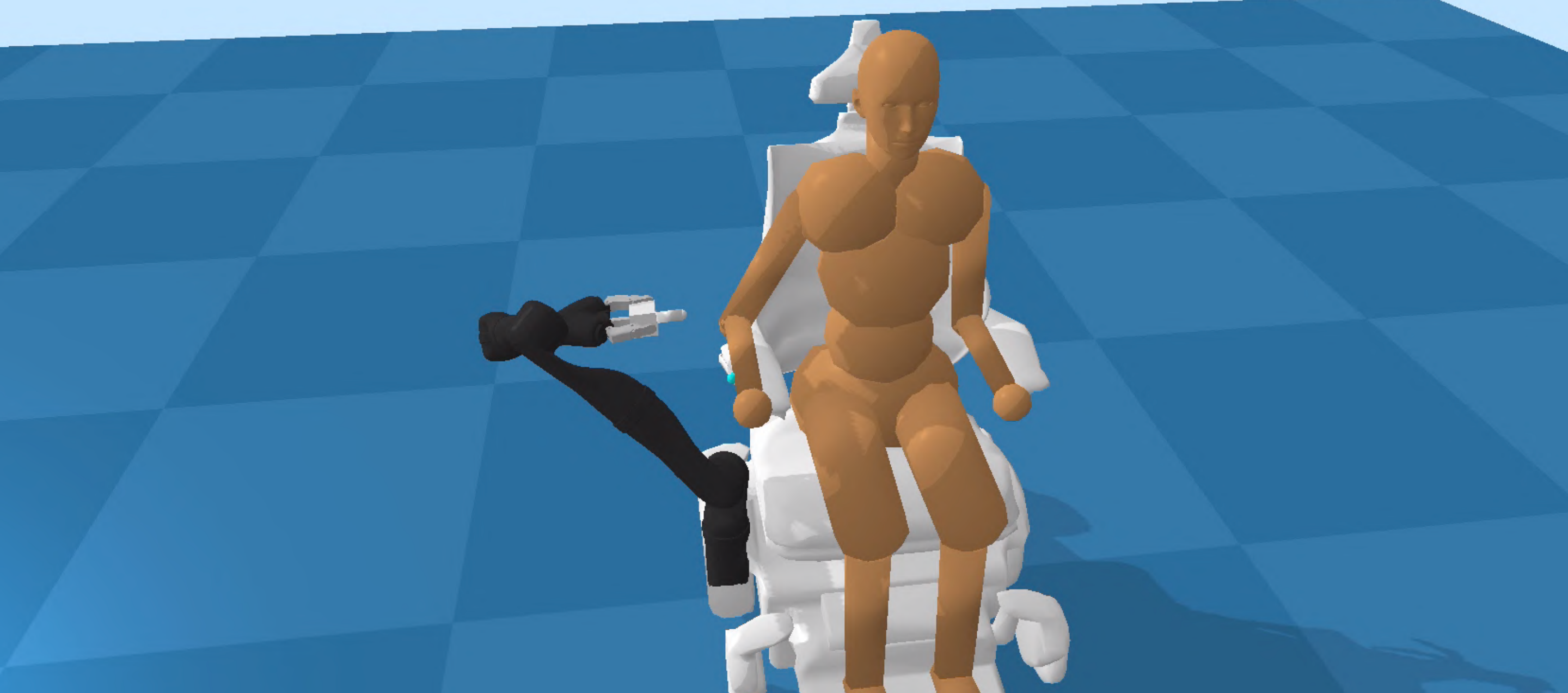}
\includegraphics[width=0.23\textwidth, trim={9cm 13cm 11cm 5cm}, clip]{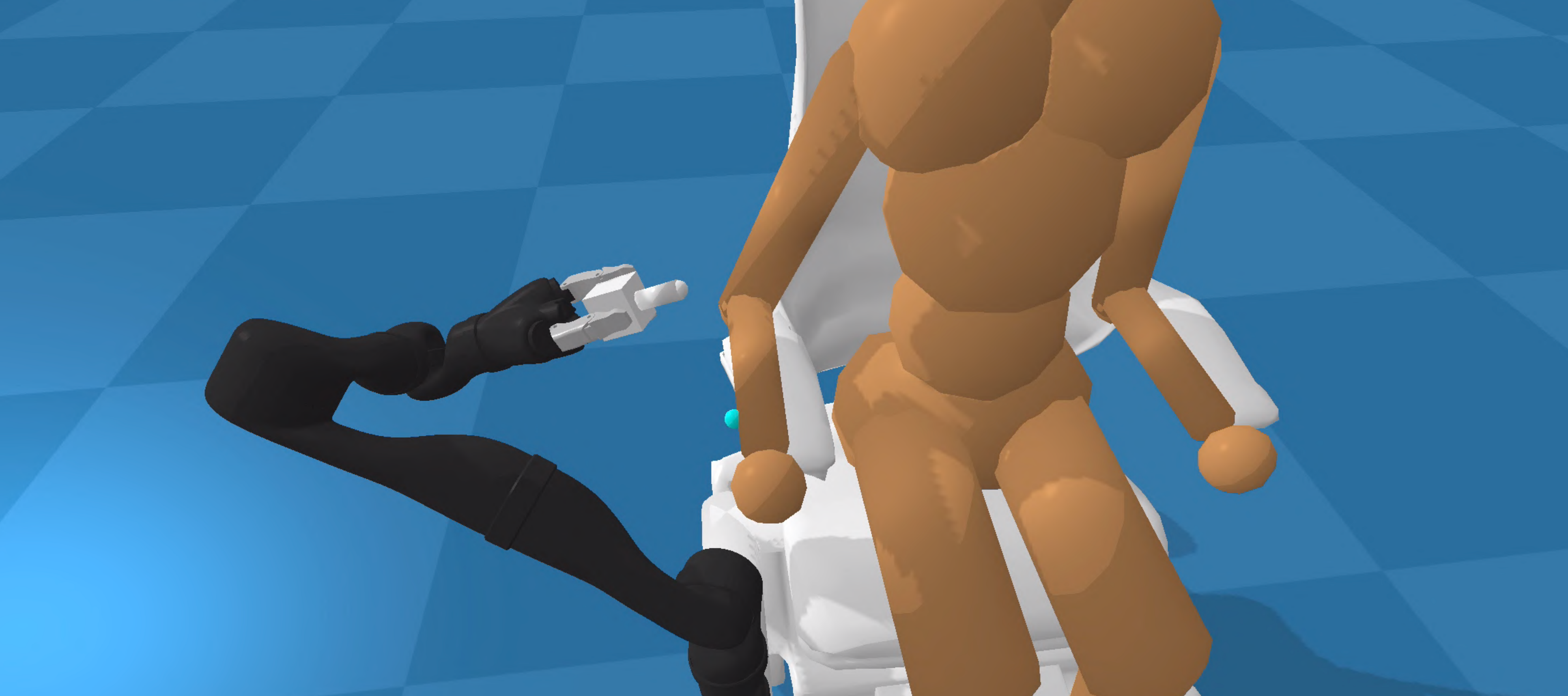}
\includegraphics[width=0.23\textwidth, trim={9cm 13cm 11cm 5cm}, clip]{scratch_itch_jaco_2}
\includegraphics[width=0.23\textwidth, trim={9cm 13cm 11cm 5cm}, clip]{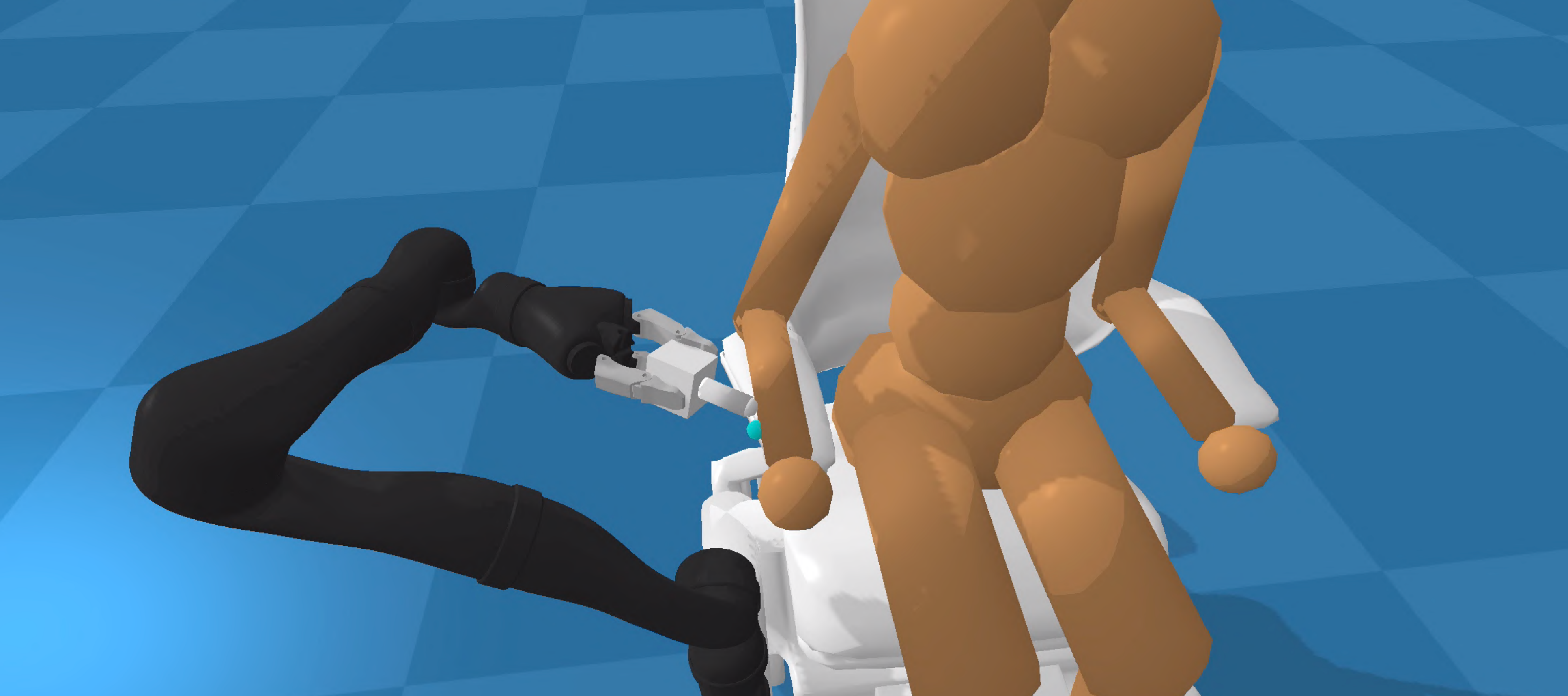}\\[0.1cm]
\raisebox{0.25in}{\rotatebox[origin=t]{90}{\parbox[][0.75cm][t]{1.8cm}{\centering Bed Bathing \\ Sawyer}}}
\includegraphics[width=0.23\textwidth, trim={9cm 8.8cm 10cm 7.5cm}, clip]{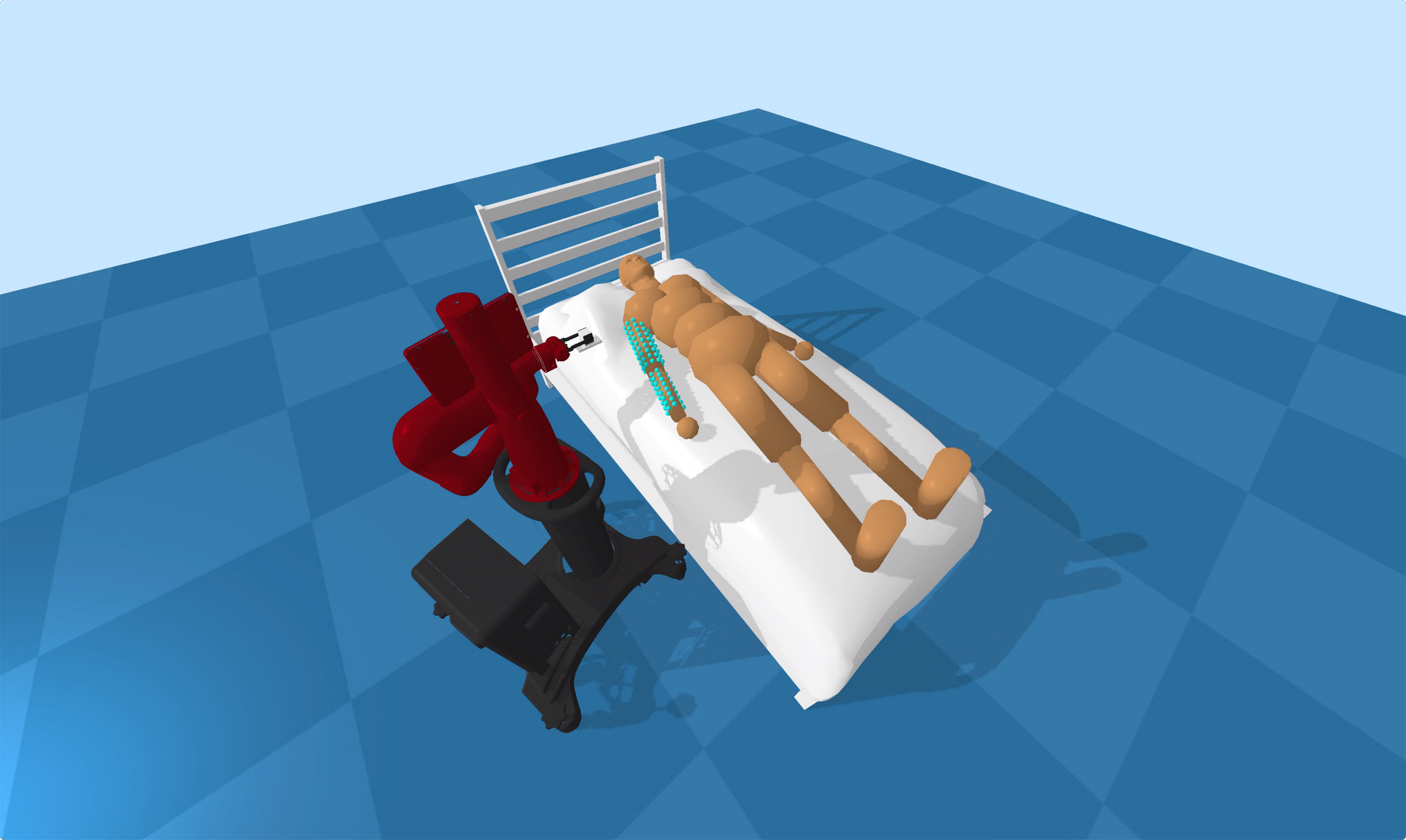}
\includegraphics[width=0.23\textwidth, trim={5cm 11cm 5cm 1cm}, clip]{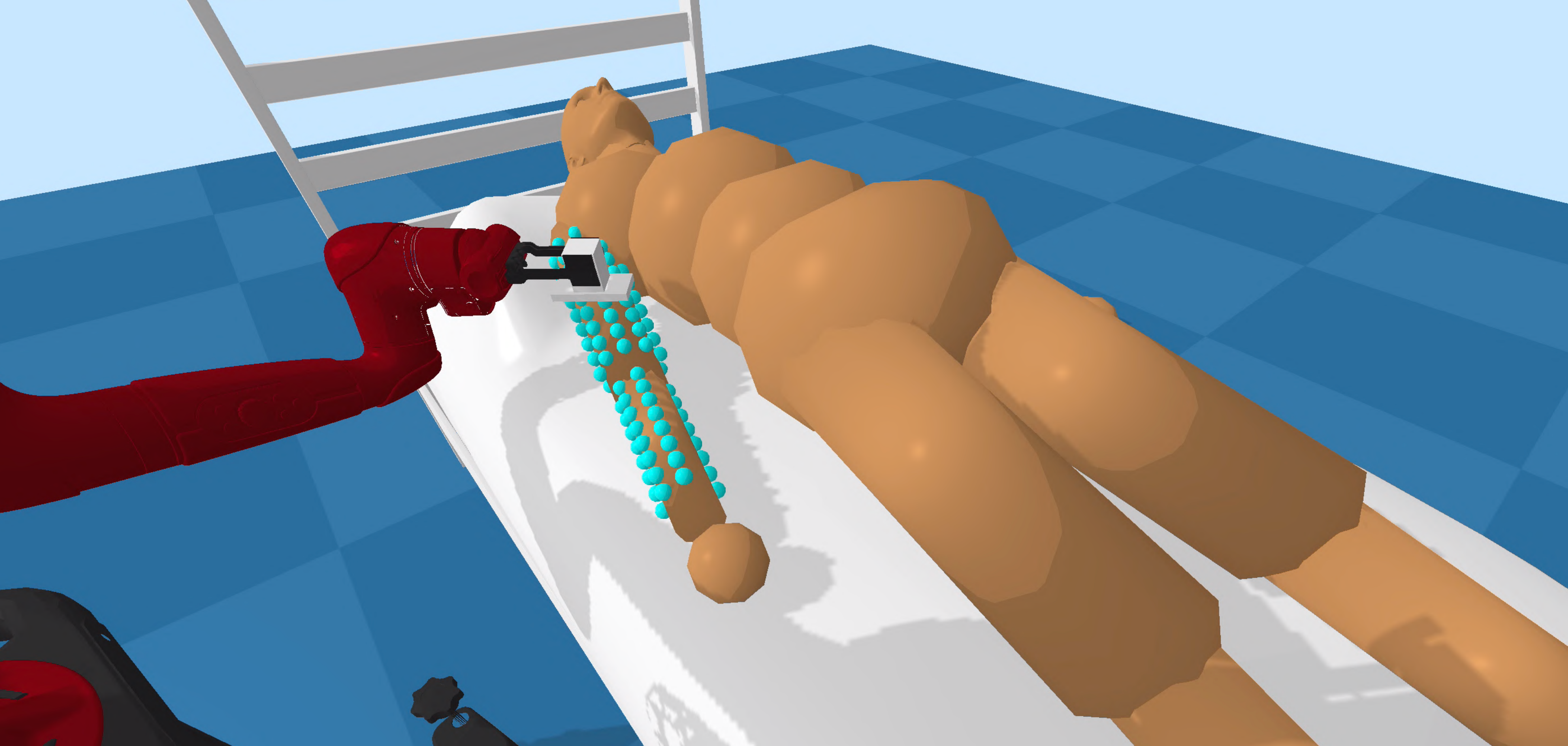}
\includegraphics[width=0.23\textwidth, trim={5cm 11cm 5cm 1cm}, clip]{bed_bathing_sawyer_2}
\includegraphics[width=0.23\textwidth, trim={5cm 11cm 5cm 1cm}, clip]{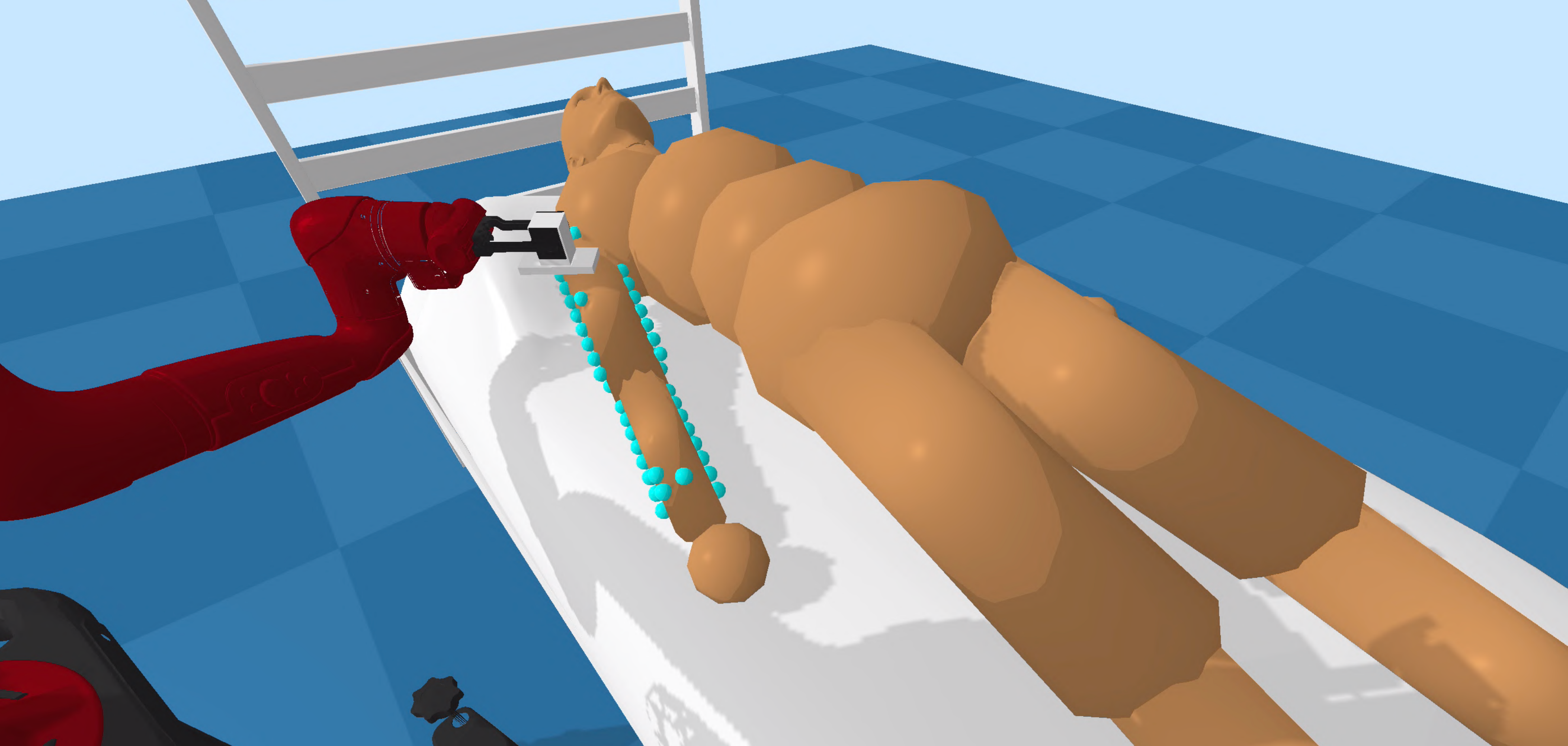}\\[0.1cm]
\raisebox{0.2in}{\rotatebox[origin=t]{90}{\parbox[][0.75cm][t]{1.4cm}{\centering Feeding \\ Baxter}}}
\includegraphics[width=0.23\textwidth, trim={10.5cm 13cm 13.25cm 7cm}, clip]{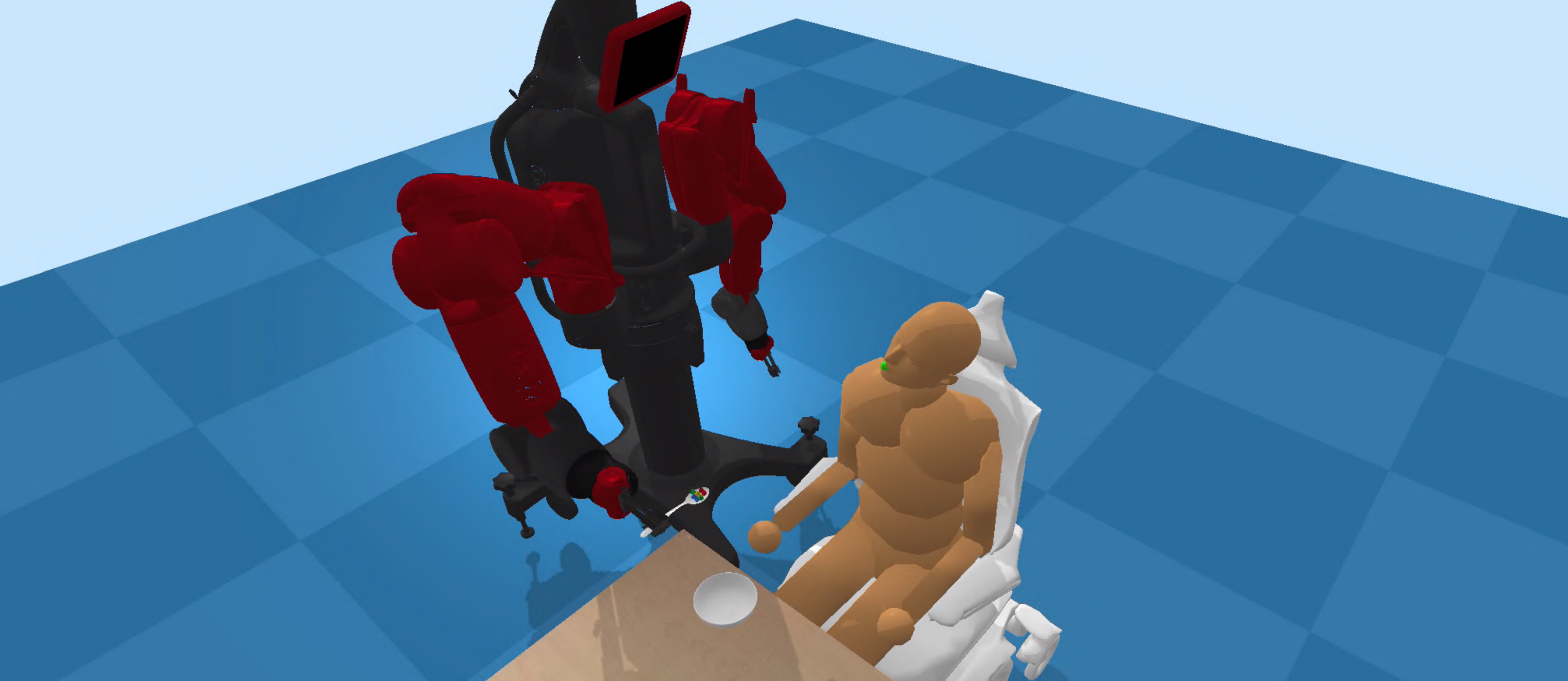}
\includegraphics[width=0.23\textwidth, trim={7cm 9cm 7cm 7cm}, clip]{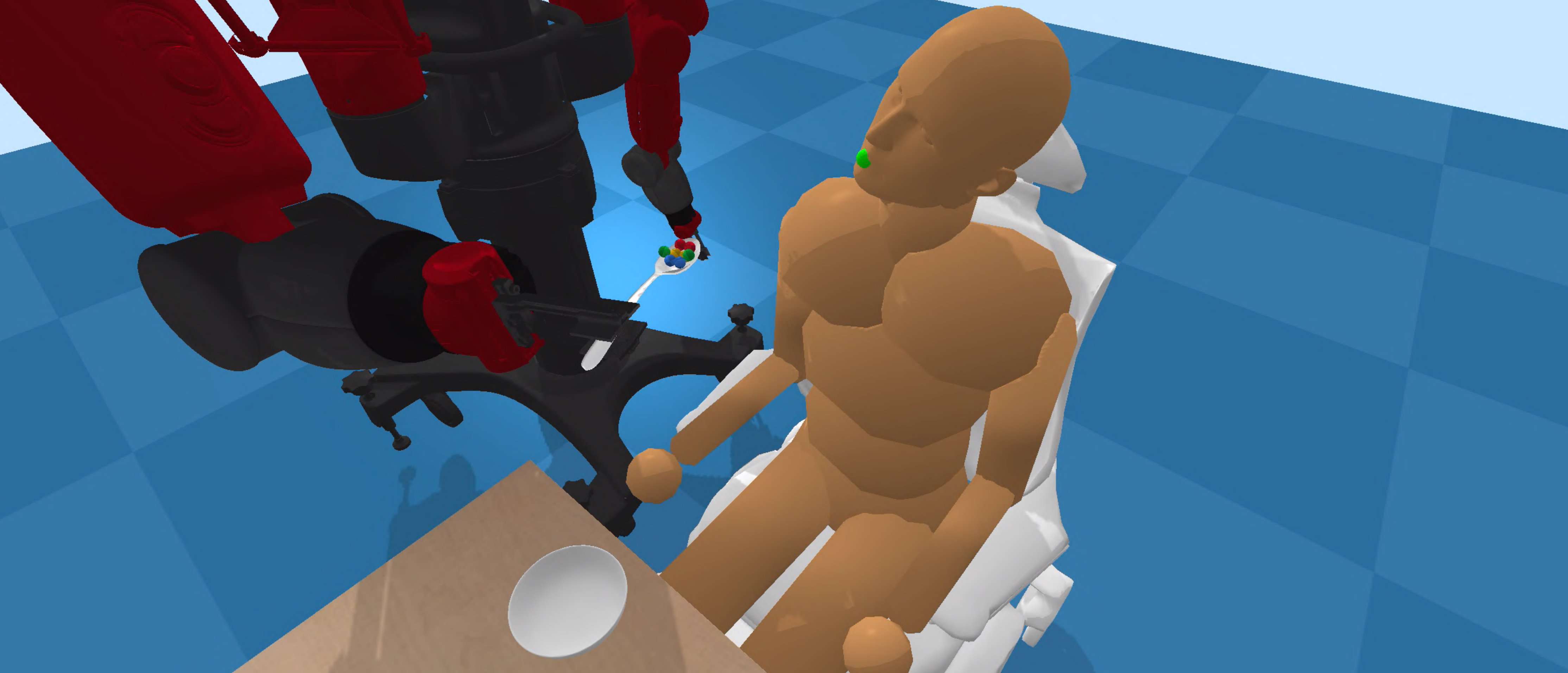}
\includegraphics[width=0.23\textwidth, trim={7cm 9cm 7cm 7cm}, clip]{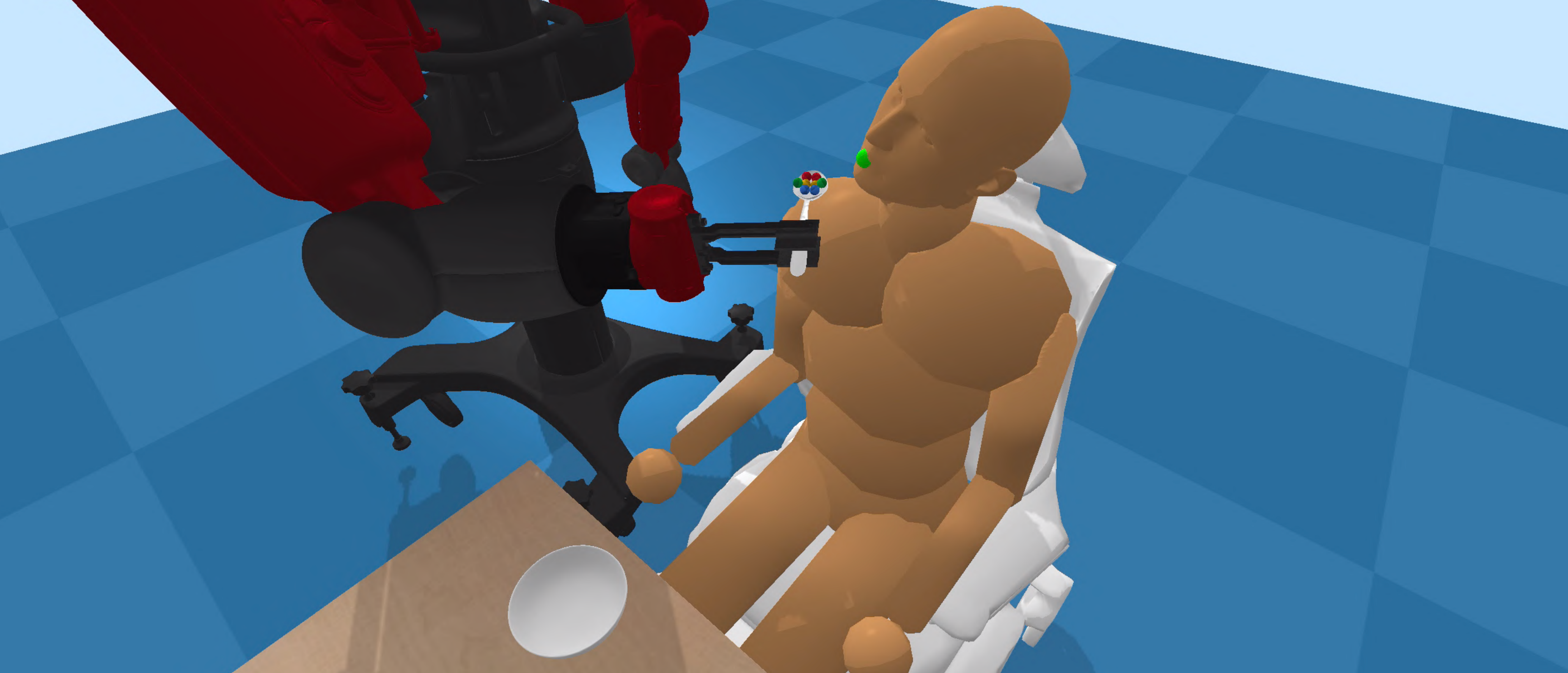}
\includegraphics[width=0.23\textwidth, trim={7cm 9cm 7cm 7cm}, clip]{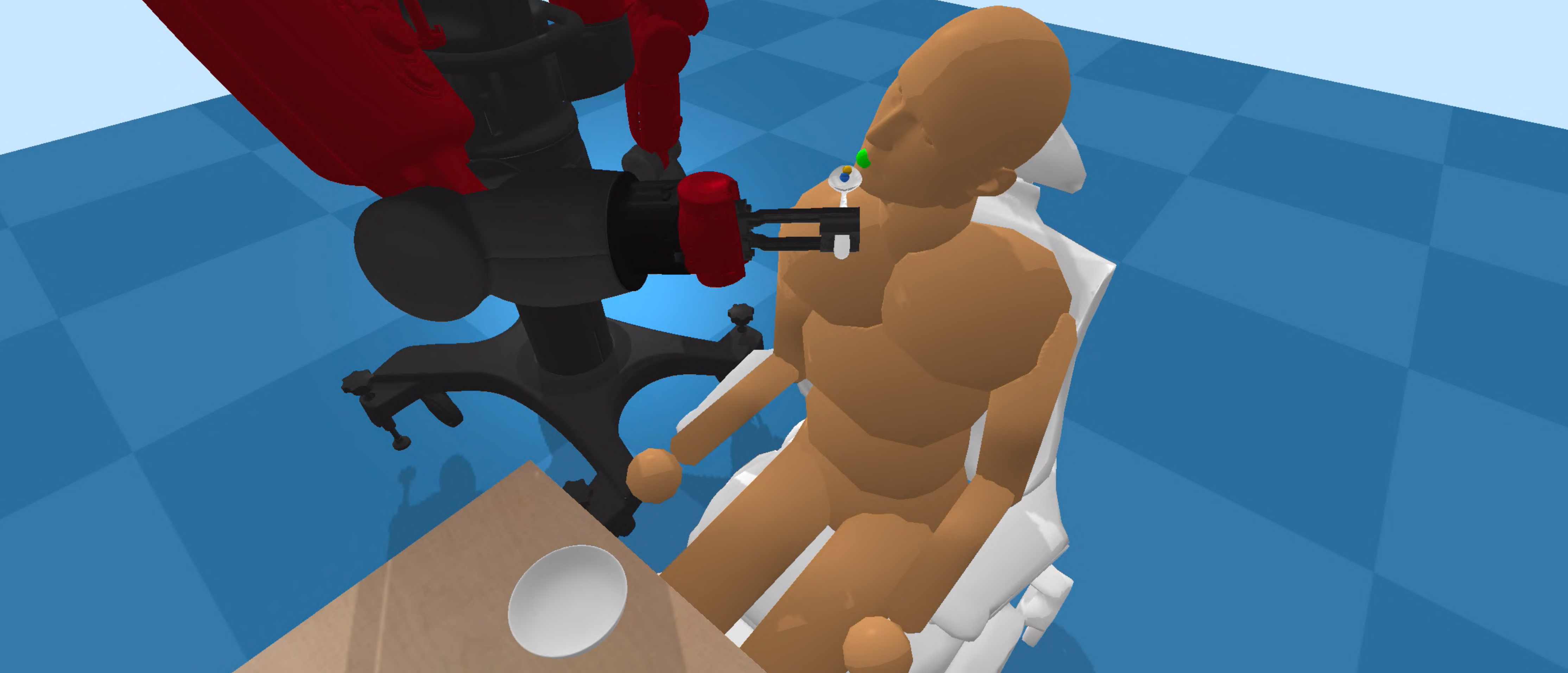}\\[0.1cm]
\raisebox{0.2in}{\rotatebox[origin=t]{90}{\parbox[][0.75cm][t]{1.4cm}{\centering Drinking \\ PR2}}}
\includegraphics[width=0.23\textwidth, trim={10.5cm 12cm 13.25cm 8cm}, clip]{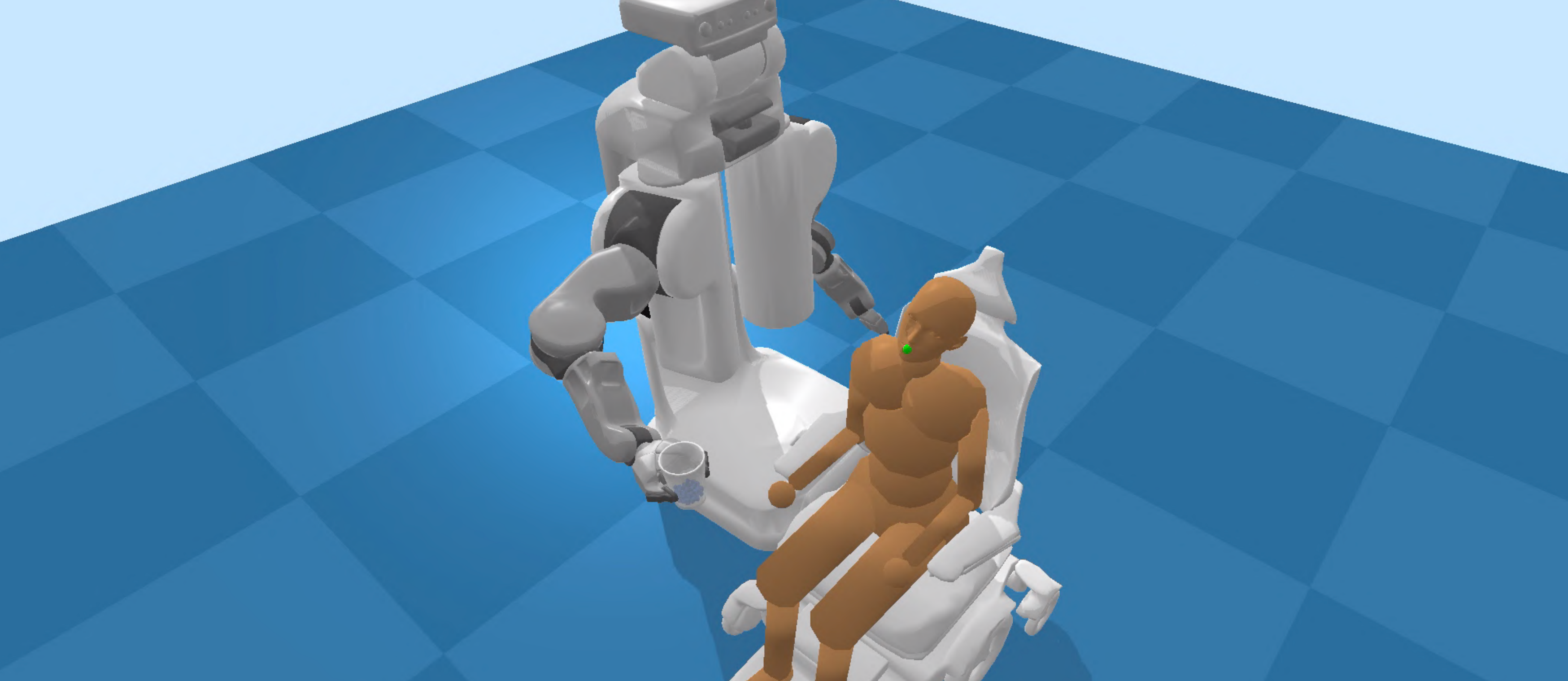}
\includegraphics[width=0.23\textwidth, trim={13.25cm 12cm 10.5cm 8cm}, clip]{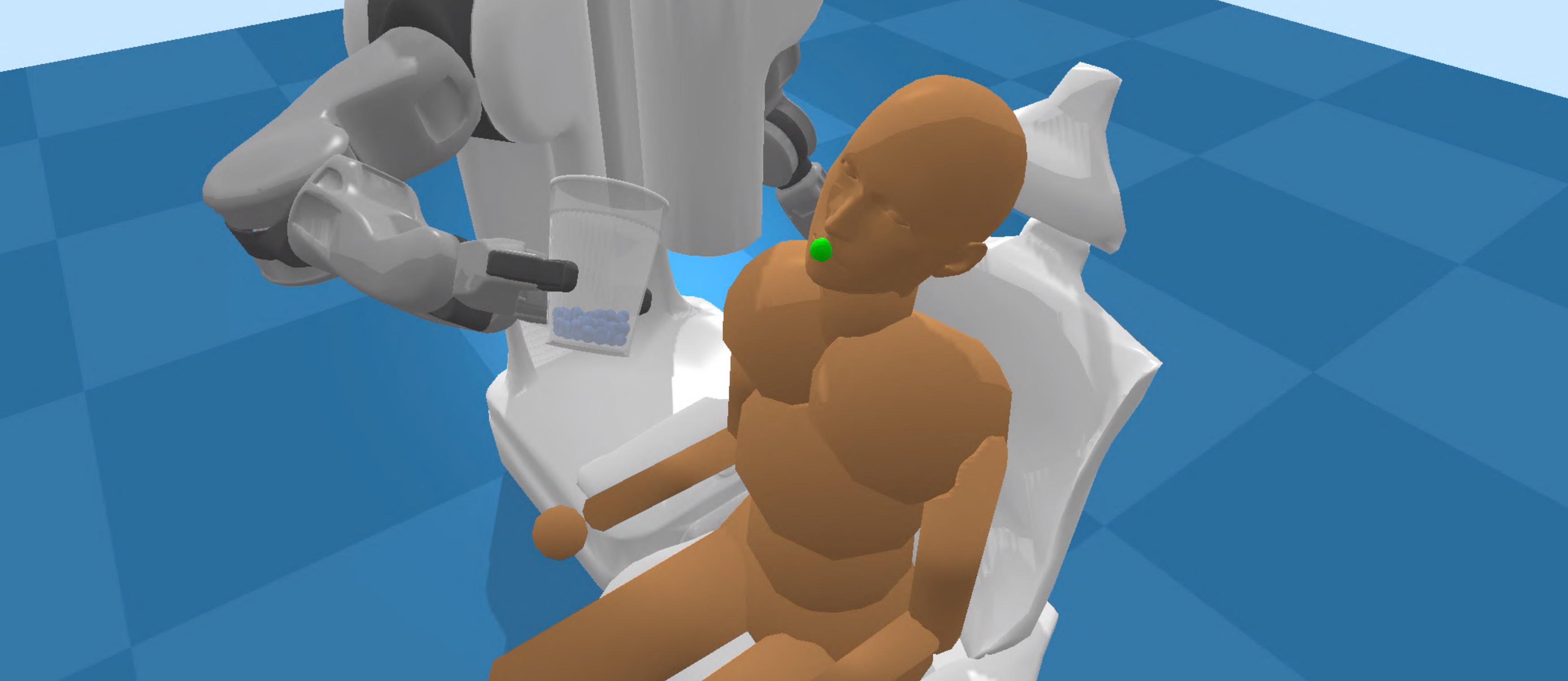}
\includegraphics[width=0.23\textwidth, trim={13.25cm 12cm 10.5cm 8cm}, clip]{drinking_pr2_2}
\includegraphics[width=0.23\textwidth, trim={13.25cm 12cm 10.5cm 8cm}, clip]{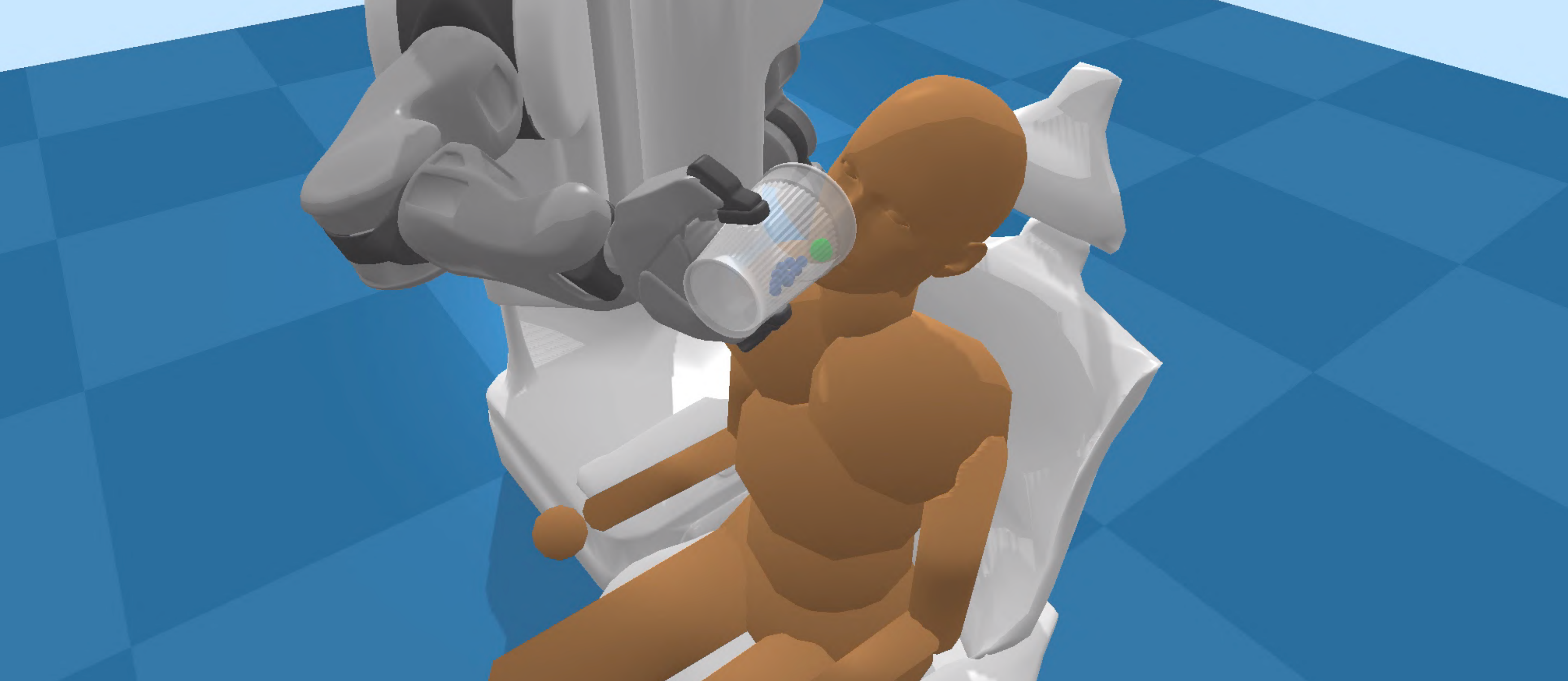}\\[0.1cm]
\raisebox{0.25in}{\rotatebox[origin=t]{90}{\parbox[][0.75cm][t]{1.75cm}{\centering Dressing \\ Baxter}}}
\includegraphics[width=0.23\textwidth, trim={10.5cm 7cm 13.5cm 11cm}, clip]{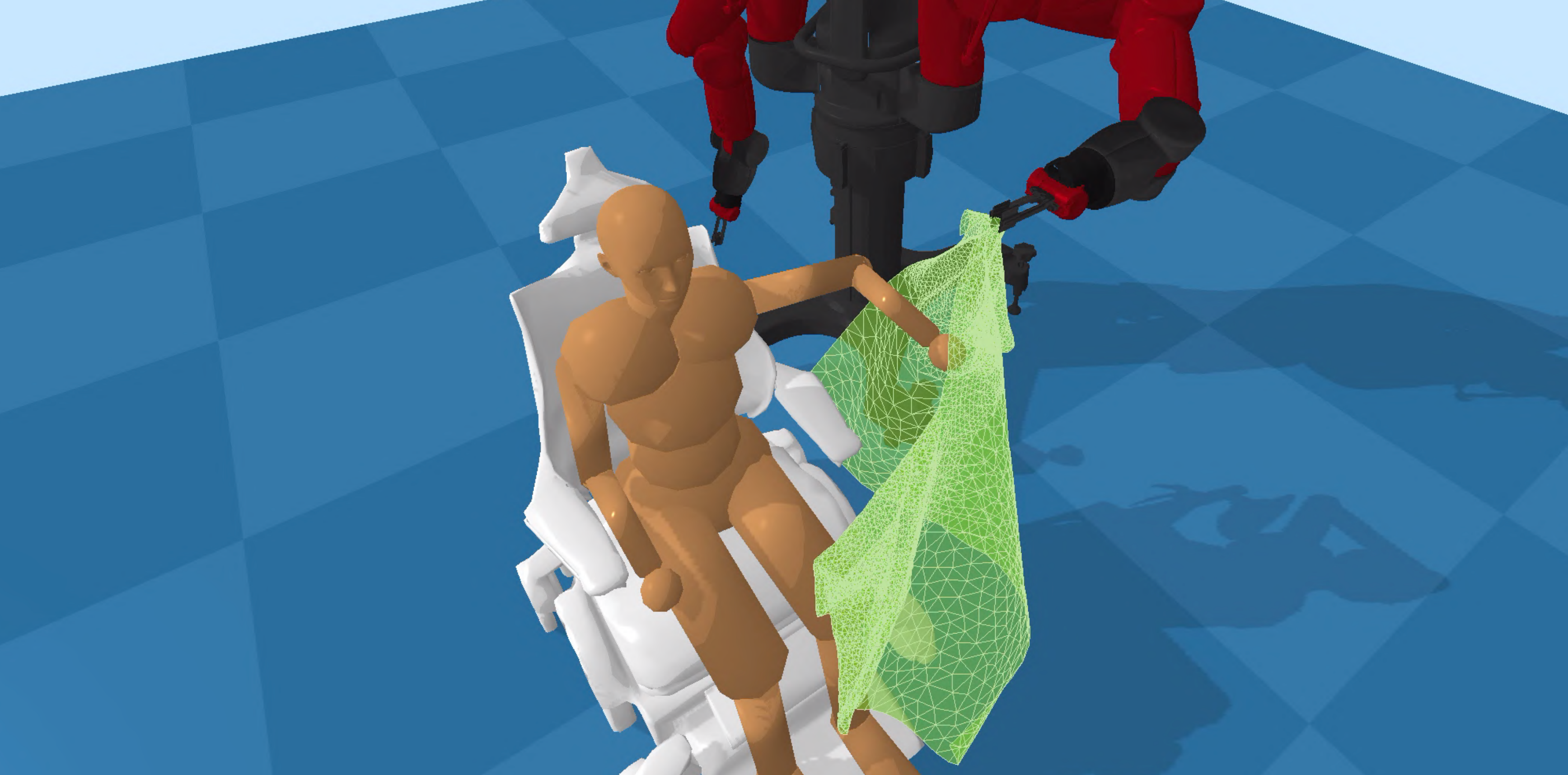}
\includegraphics[width=0.23\textwidth, trim={6.5cm 8cm 14cm 8cm}, clip]{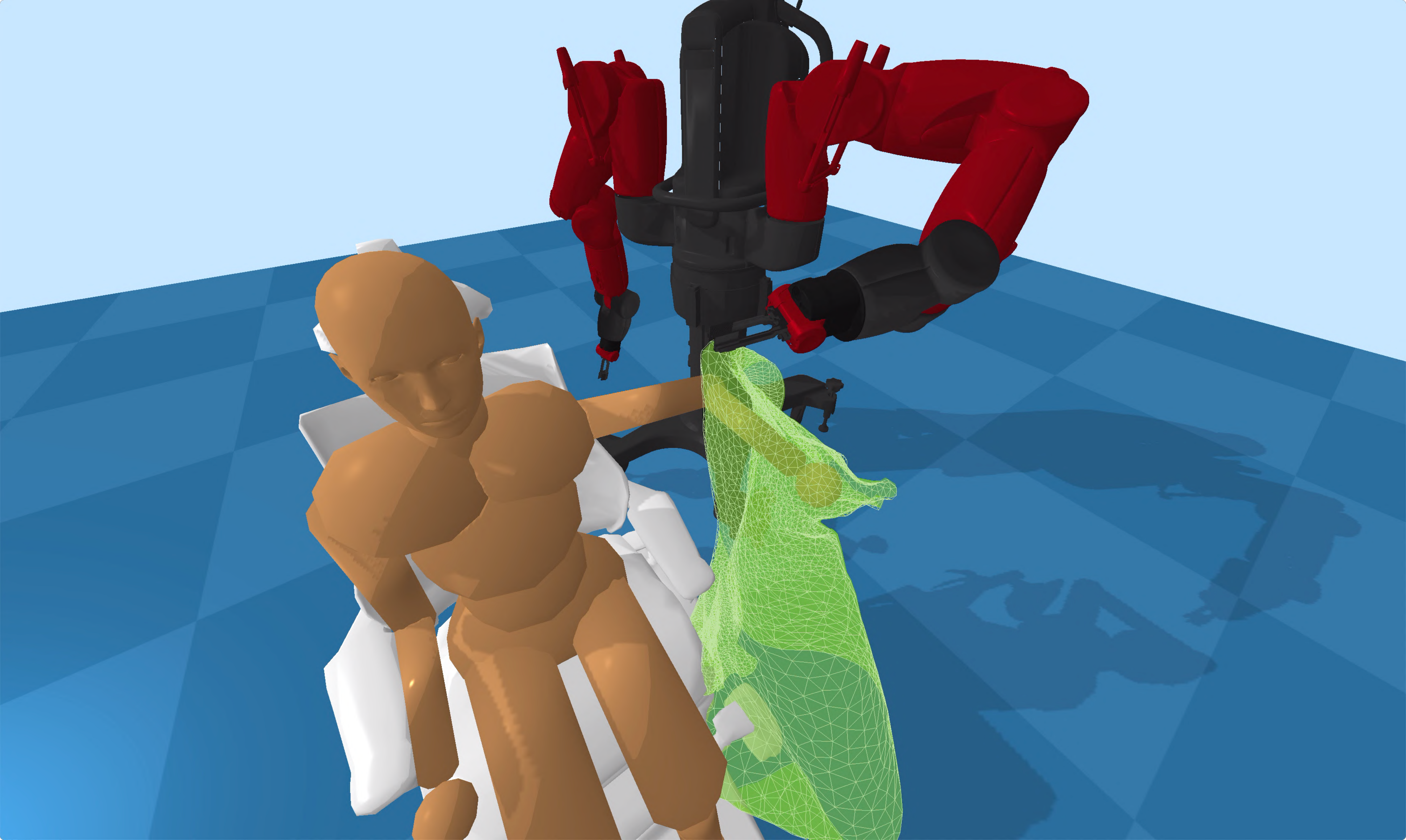}
\includegraphics[width=0.23\textwidth, trim={6.5cm 8cm 14cm 8cm}, clip]{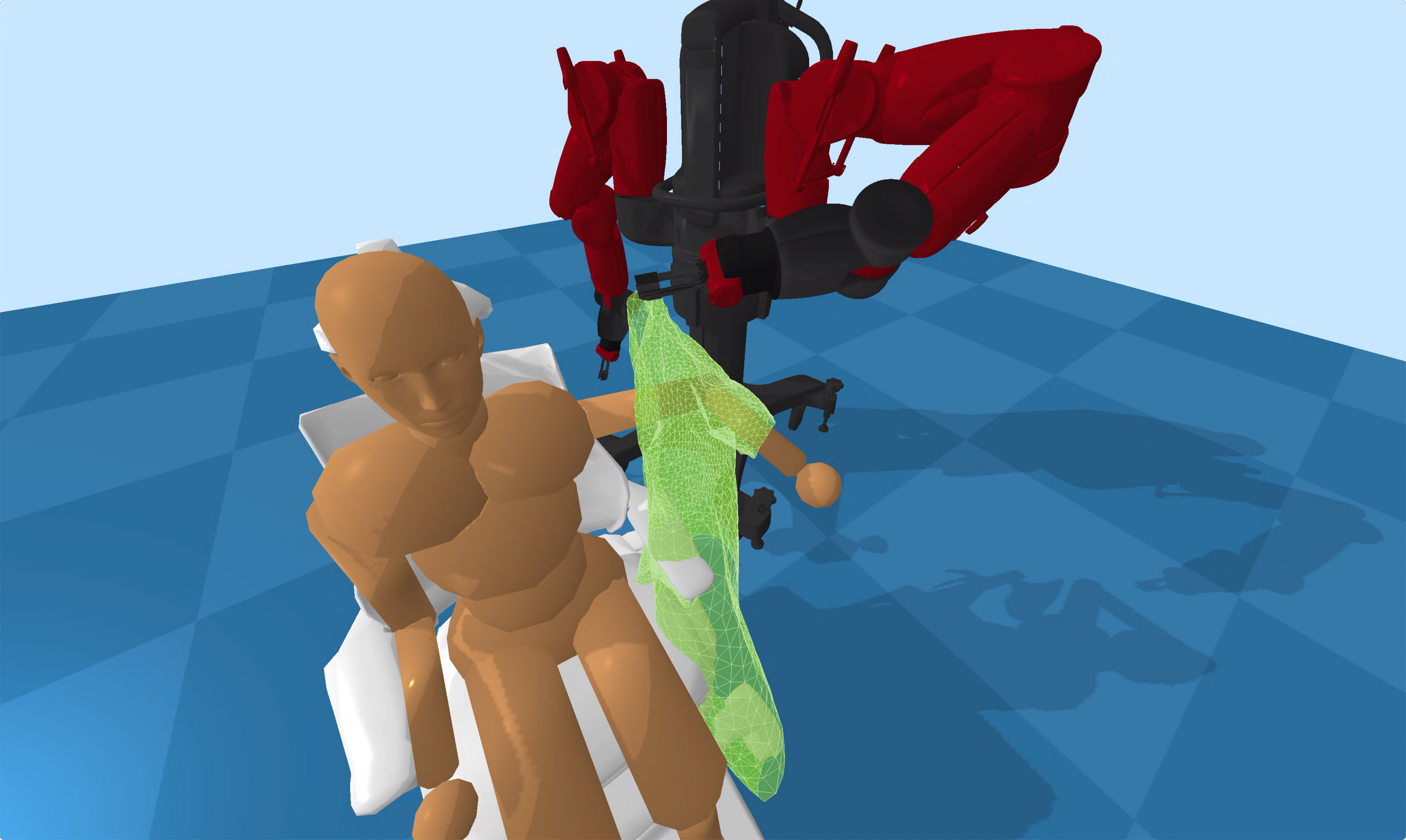}
\includegraphics[width=0.23\textwidth, trim={6.5cm 8cm 14cm 8cm}, clip]{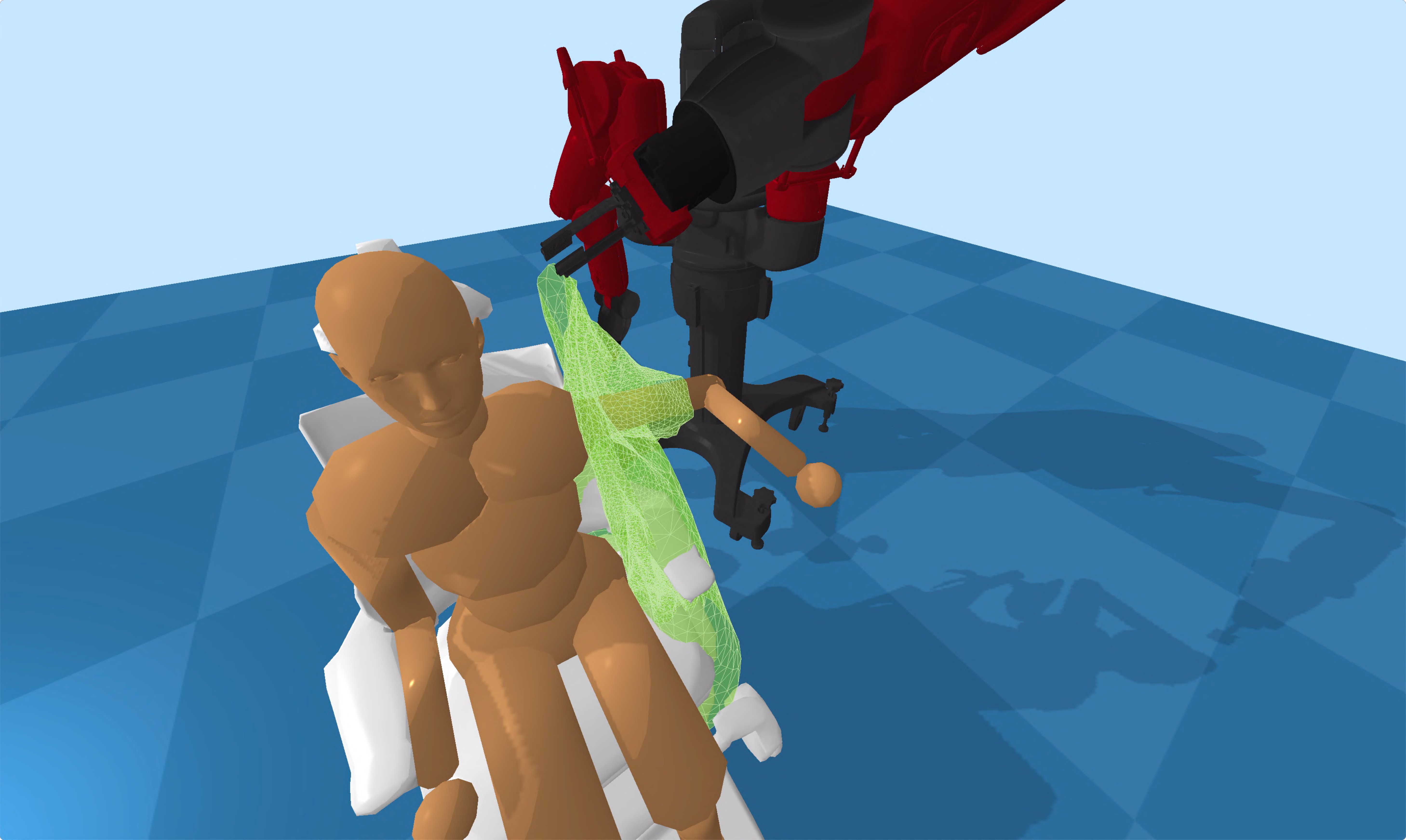}\\[0.1cm]
\ \raisebox{0.25in}{\rotatebox[origin=t]{90}{\parbox[][0.75cm][t]{1.9cm}{\centering Arm Manip. \\ PR2}}}
\includegraphics[width=0.23\textwidth, trim={5cm 7.5cm 14cm 7.25cm}, clip]{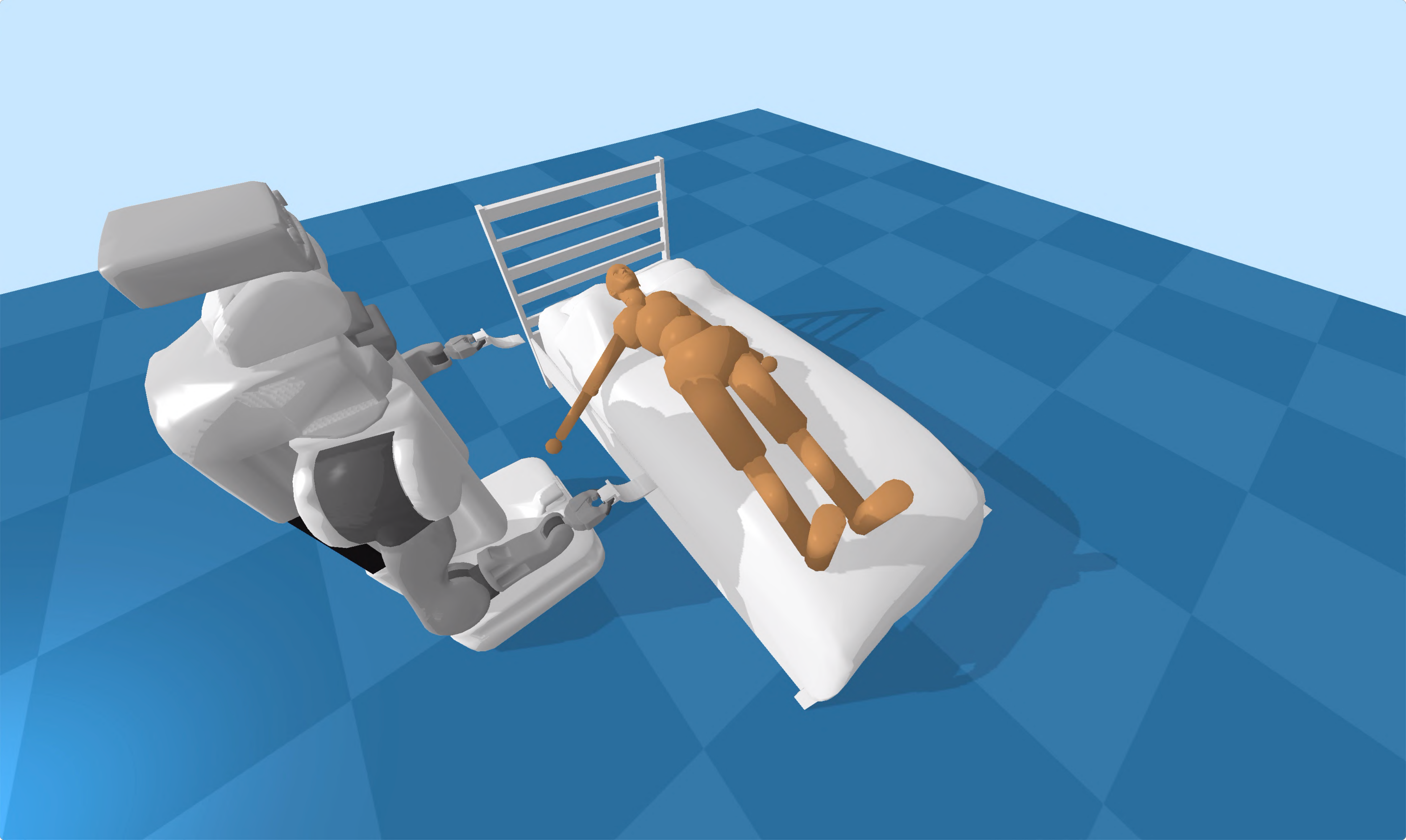}
\includegraphics[width=0.23\textwidth, trim={2cm 7cm 8cm 3cm}, clip]{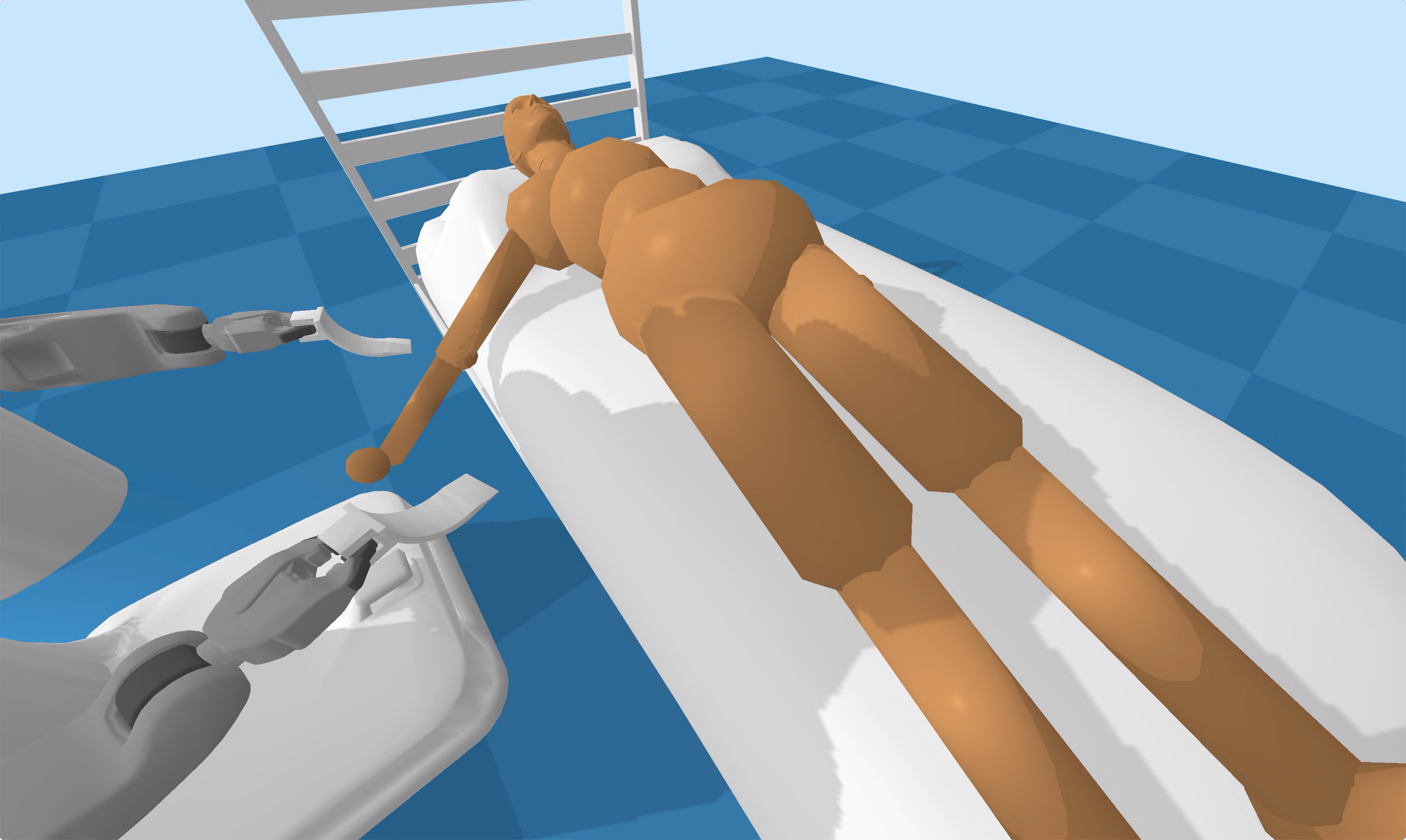}
\includegraphics[width=0.23\textwidth, trim={2cm 7cm 8cm 3cm}, clip]{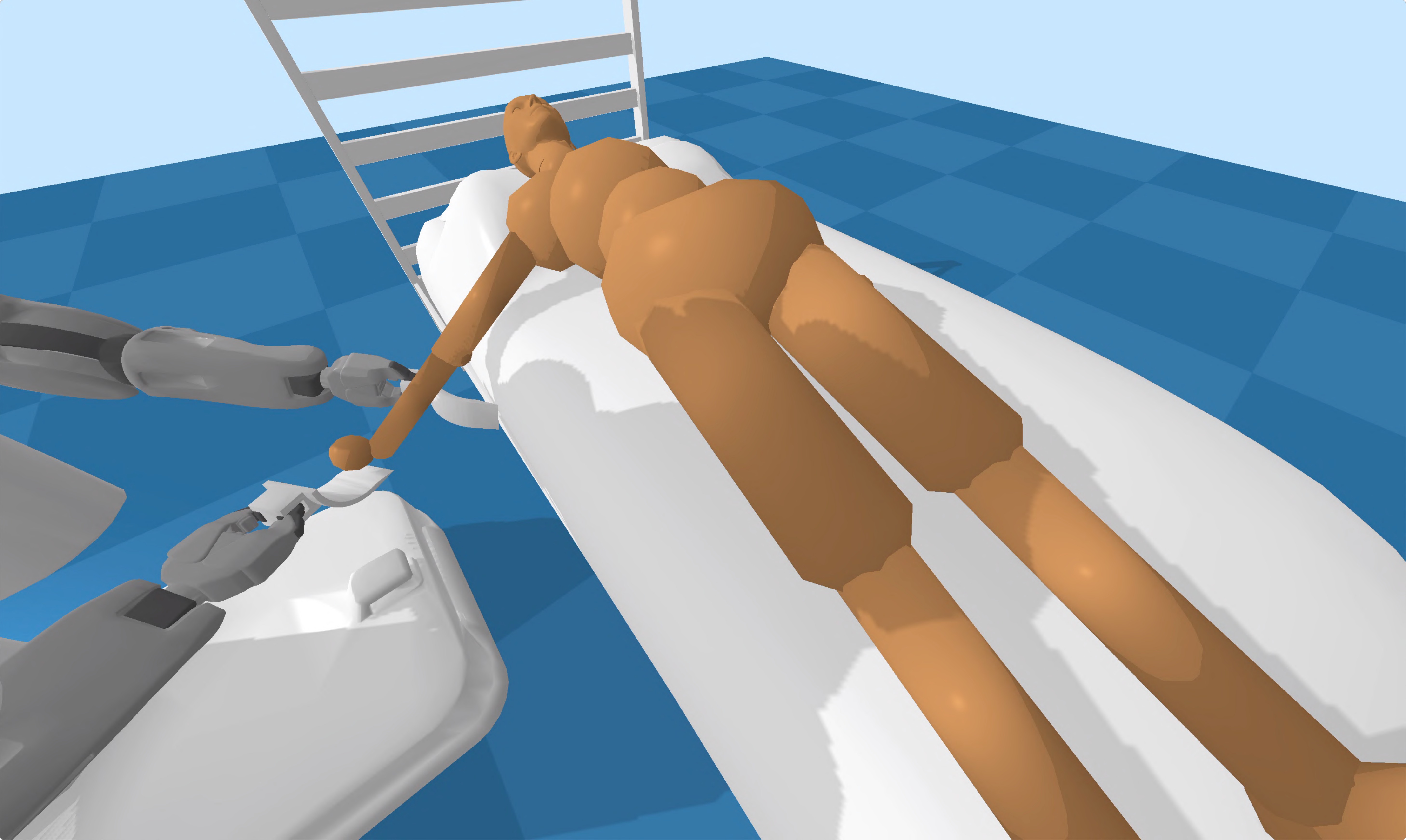}
\includegraphics[width=0.23\textwidth, trim={2cm 7cm 8cm 3cm}, clip]{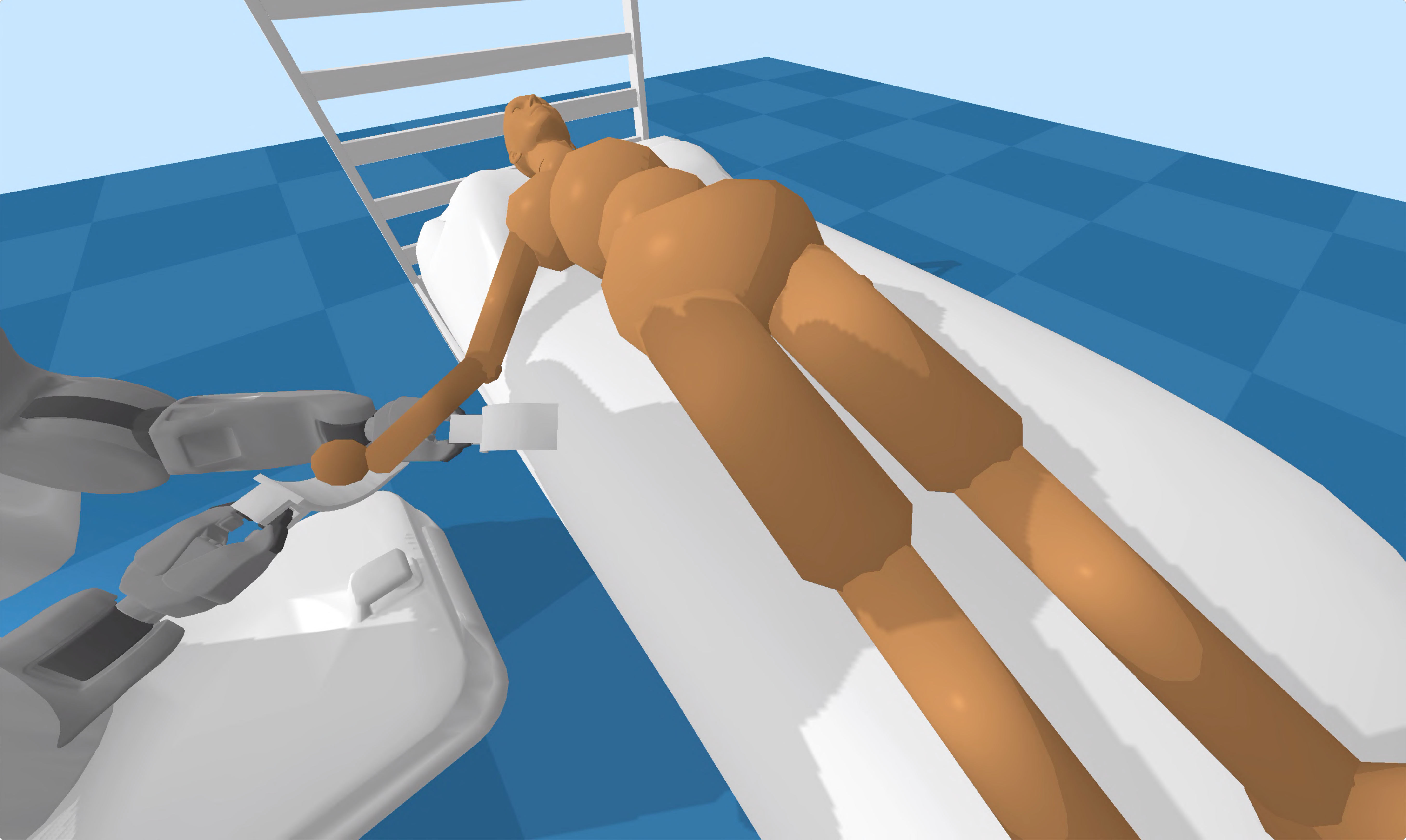}
\vspace{-0.2cm}
\caption{\label{fig:sequences}Image sequences from executing trained robot policies for each of the six assistive tasks when the person holds a static pose.}
\vspace{-0.4cm}
\end{figure*}

\section{Policy Learning and Control}
\label{sec:control}

Assistive Gym supports position control for commanding the robot. Actions for each robot's 7-DoF arm are represented as changes in joint positions, $\Delta P \in \mathbb{R}^7$ for single robot arms or $\Delta P \in \mathbb{R}^{14}$ for dual arm robots. Actions for a person include $\Delta P \in \mathbb{R}^{10}$ for the human arm, or $\Delta P \in \mathbb{R}^4$ for the head. 
We limit the strength of each robot actuator to reduce the likelihood of a robot learning policies that apply high forces to the person.
This limitation could potentially be removed in future iterations using techniques, such as curriculum learning, to balance between making task success and satisfying human preferences~\cite{cleggdissertation}.
In addition, prior research has shown that real robots can provide assistance based on what they have learned in simulation~\cite{kapusta2019dressing, erickson2018deep}, yet further work will be needed to enable real robots to benefit from controllers learned in Assistive Gym.

At each time step, the robot records observations from the state of the system, executes an action according to a control policy, and then receives a reward.
We give observations to the robot in accordance to observations that can be obtained in a real-world assistive robotics scenario. This includes, the 3D position and orientation of the robot's end effector, the 7D joint positions of the robot's arm, forces applied at the robot's end effector, and 3D positions of task relevant joints along the human body, such as the wrist, elbow, shoulder, or the position and orientation of a person's head. Joint positions of a real human body can be obtained using a number of existing approaches, such as with a single image using OpenPose, or the pose of a person on a bed using a pressure sensing mat~\cite{cao2018openpose, clever20183d}. All positions are defined with respect to the robot's torso, or base position.

In this work, we use deep reinforcement learning technique, proximal policy optimization (PPO), to learn control policies for robotic assistance.
PPO is a policy gradient algorithm used across a number of contexts, from Atari games to real-world quadruped robot locomotion~\cite{schulman2017proximal, tan2018sim}.
We follow the original policy representation as presented in~\cite{schulman2017proximal}, using a fully-connected neural network with two hidden layers of 64 nodes, and tanh activations.
When performing co-optimization in Section~\ref{sec:collaborative}, we train two policies concurrently, for the robot and active human, with both optimized using a shared reward, $r(\bm{s})$ (Section~\ref{sec:prefs}).

\section{Evaluation}
\label{sec:evaluation}

In the following sections, we present and analyze baseline control policies for the four robots assisting with the six assistive tasks. For each task, we train robot controllers using PPO, as described in Section~\ref{sec:control}. We train each policy using 36 concurrent simulation actors for a total of 10,000,000 time steps, or 50,000 simulation rollouts (trials). Each simulation rollout consists of 200 time steps (20 seconds of simulation time at 10 time steps per second), where a policy can execute a new action at each time step. We perform a 10 epoch update of the policy after each actor completes a single rollout (i.e. every 7,200 time steps). We trained all policies using Amazon Web Services (AWS) with a 36 vCPU machine. 
Training times varied from 2 hours (itch scratching) to 19 hours (drinking). Training a policy for dressing assistance took $\sim$6 days due to simulating dynamic cloth.

\begin{figure*}
\centering
\includegraphics[width=0.24\textwidth, trim={10cm 13cm 10cm 5cm}, clip]{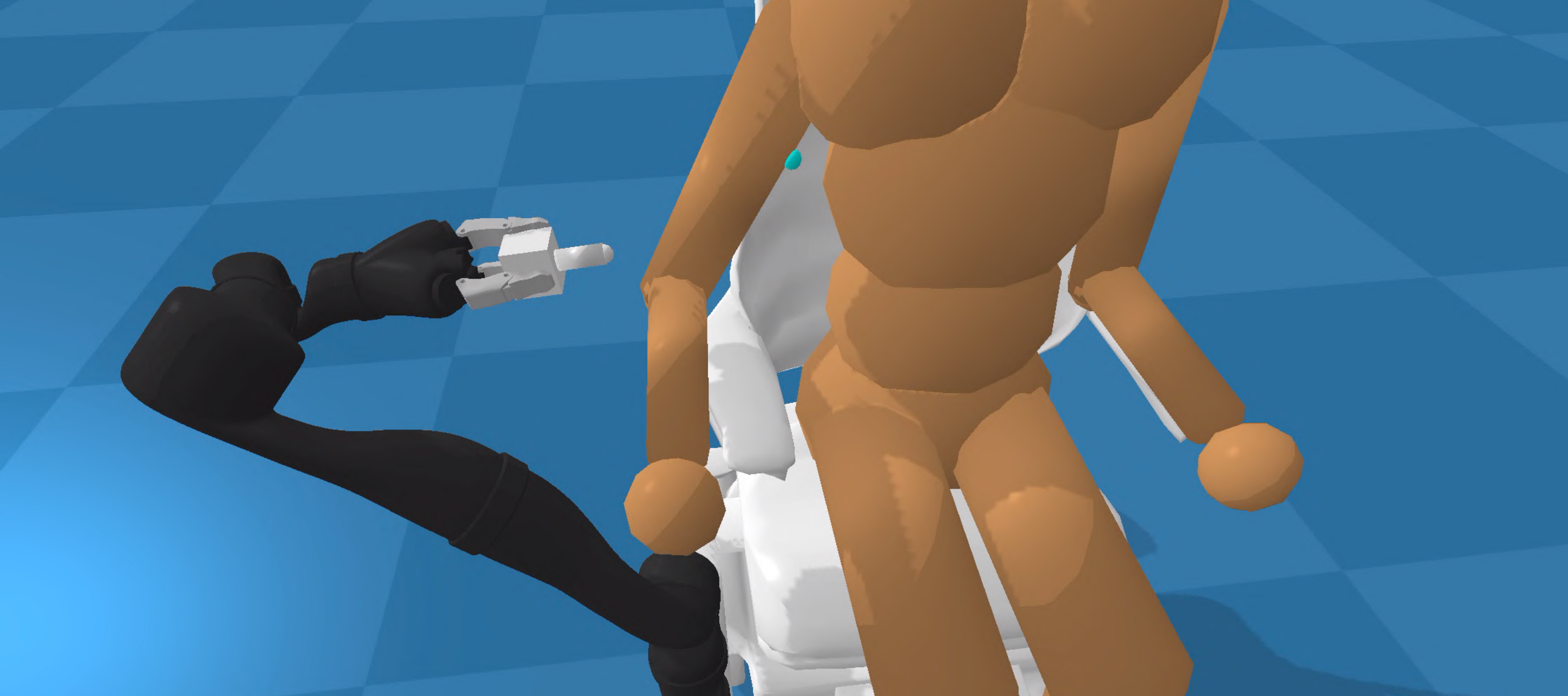}
\includegraphics[width=0.24\textwidth, trim={10cm 13cm 10cm 5cm}, clip]{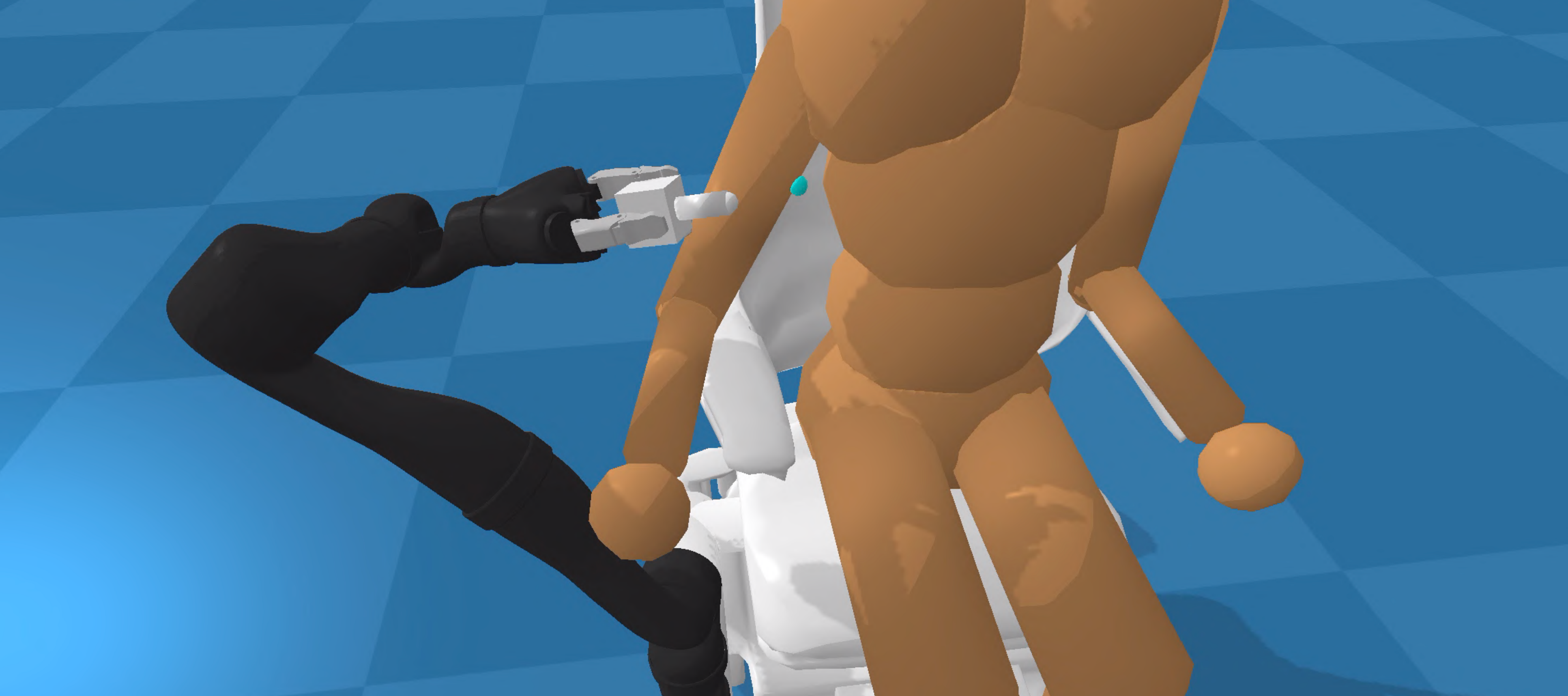}
\includegraphics[width=0.24\textwidth, trim={10cm 13cm 10cm 5cm}, clip]{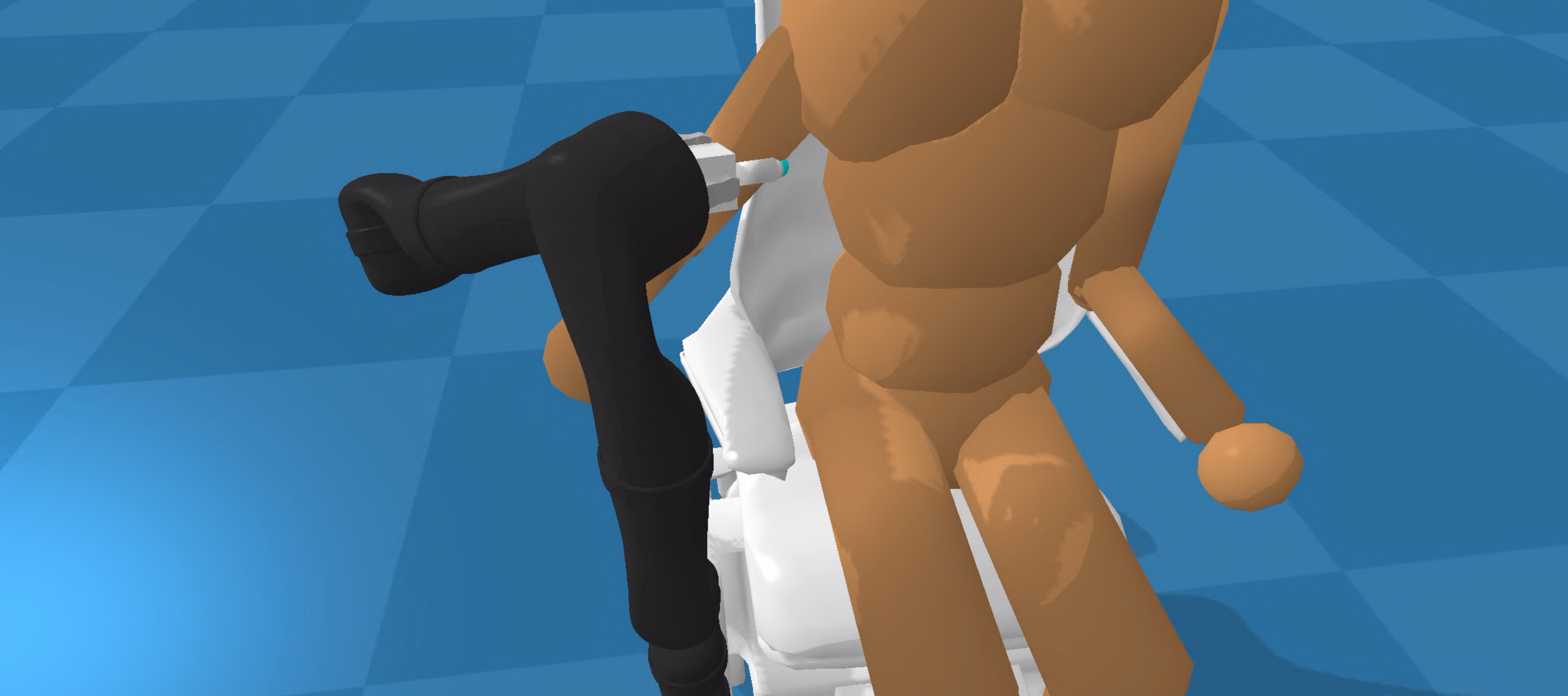}
\includegraphics[width=0.24\textwidth, trim={10cm 13cm 10cm 5cm}, clip]{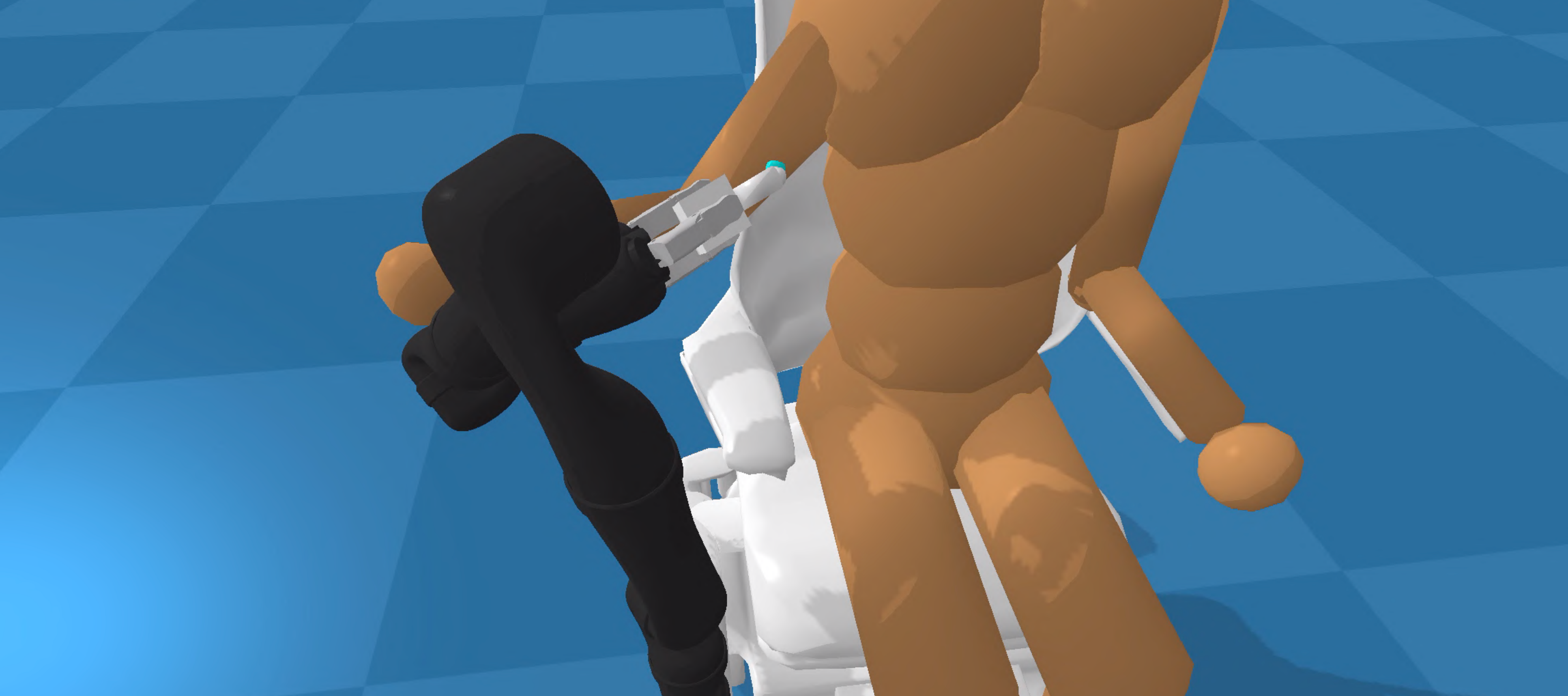}
\vspace{-0.2cm}
\caption{\label{fig:scratching_coop}Image sequence of a human policy rotating the person's arm so that the Jaco robot can better scratch an itch underneath their arm.}
\vspace{-0.4cm}
\end{figure*}

\subsection{Robotic Assistance}

Similar to prior assistive robotic studies, we consider the scenario where a robot provides active physical assistance to a person who attempts to hold a static body pose~\cite{erickson2018deep, chen2014investigation}. 

For each robot, we used PPO to train a separate control policy for each of the six assistive tasks.
This amounts to a total of 6$\times$4 = 24 unique policies.
We then repeat each policy training with 3 random seeds and select the best of the three policies, based on the average reward achieved over the last 10 policy updates (the last 360 simulation rollouts).

Fig.~\ref{fig:sequences} shows image sequences for select robots executing their learned policies over all six of the assistive tasks.
We note that for the first four tasks (itch scratching, bed bathing, feeding, and drinking), PPO was able to learn reasonable control policies for all four robots, with varying levels of performance between the robots.
However, both the dressing and arm manipulation tasks remain challenging for all four of the robots. For dressing assistance, the robots struggled to pull the hospital gown up a person's forearm and upper arm, yet could on occasion when the opening of the gown sleeve was randomly placed close to the person's fist, as seen in Fig.~\ref{fig:sequences}. For arm manipulation assistance, policy optimization had difficulties learning to lift a person's arm back onto the bed while also adhering to the person's preferences. The trained policies often had the robot use the thin edge of the tool to lift the person's arm, resulting in high pressure applied to the person and subsequently a large negative reward.

Assistive Gym also provides an opportunity to compare various robot platforms in terms of their ability to provide physical assistance to people. 
To compare robots, we held all parameters and settings for PPO and the simulation environments constant.
Given a trained control policy for a specific robot and assistive task, we evaluated the policy over 100 simulation rollouts of the task. Table~\ref{table:robot_comparison} lists the average reward each robot achieved over the 100 simulation rollouts for each task. Assistive Gym also defines task success for each task, and we include success rates for the best performing robot in Table~\ref{table:robot_comparison}.

Overall, we observed that there are significant opportunities for improvement for all of the robots when providing assistance to a static human across the six tasks.
In addition, these results can help compare the physical capabilities of each robot. For example, the PR2's shorter arm span made some tasks more difficult, such as itch scratching, which often requires reaching around the person's arm.

\begin{table}
\centering
\vspace{6pt}
\caption{\label{table:robot_comparison}Average reward for 100 trials with a \textbf{static human}. Task success on 100 trials for the robot with the highest reward.}
\begin{tabular}{cccccc} \toprule
    Task & PR2 & Jaco & Baxter & Sawyer & \textit{Success} \\ \midrule\midrule
    Itch Scratching & 55.1 & \textbf{280.8} & 225.4 & 136.8 & \textbf{54\%} \\
    Bed Bathing & 86.7 & 104.4 & 88.4 & \textbf{109.0} & \textbf{24\%} \\
    Feeding & 100.5 & 83.8 & \textbf{108.5} & 95.6 & \textbf{88\%} \\
    Drinking & 182.5 & 85.7 & 263.3 & \textbf{436.0} & \textbf{72\%} \\
    Dressing & \textbf{11.5} & -17.0 & 5.6 & -27.6 & \textbf{26\%} \\
    Arm Manipulation & \textbf{-162.4} & -177.5 & -228.1 & -210.6 & \textbf{8\%} \\
	\bottomrule
\end{tabular}
\vspace{-0.4cm}
\end{table}

\subsection{Collaborative Assistance}
\label{sec:collaborative}

In the prior sections, we demonstrated how robots can learn to assist a person while the person holds a static pose. However, there are often scenarios in which a person who requires assistance will have some limited motor functionality. For example, a person may have a full range of motion of their arms, but suffer from large tremors, or conversely, may have fine motor control, but suffer from muscle weakness or a limited range of motion. In these cases, a person may prefer to actively help the robot accomplish its task, rather than hold a static pose. This is in part due to the assumption that people will be collaborative while receiving assistance, as they directly benefit from the assistance a robot provides.

We model human motion using co-optimization, in which both the human and robot are active agents that are trained simultaneously. We train separate control policies for the human and robot using PPO. Both the robot and human share the same reward function, but have different observation and action sets. For example, the observations for the robot include the robot's joint angles, whereas the observations for the human include the human's joint angles (proprioception). 

This is inspired by work from Clegg et al., who explored co-optimization between a simulated KUKA IIWA robot and active human for robot-assisted dressing~\cite{cleggdissertation, clegg2020ral}. Their work found that co-optimization can lead to more realistic human motions and improved assistance from the robot.
Building upon their initial findings, we apply co-optimization across a wider variety of robots and assistive tasks.
We give the person's arm and head just enough motor strength to lift up against gravity.
Similar to the previous evaluations, we hold all parameters and settings for PPO and the simulation environments constant between the four different robots. With co-optimization, we retrain policies for all of the robots and tasks, again choosing the best trained policy from three random seeds. Each robot policy has an associated human policy learned during co-optimization. We then evaluated each trained human and robot policy over 100 simulation rollouts for a specific task. Table~\ref{table:human_collaboration} depicts the average reward each robot achieved over the 100 simulation rollouts for each task when assisting an active human. Note that we do not evaluate the arm manipulation task with human motion, since an active human is often able to lift their arm back onto the bed, unaided by the robot.

\begin{table}
\centering
\vspace{6pt}
\caption{\label{table:human_collaboration}Average reward for 100 trials with an \textbf{active human}. Task success on 100 trials for the robot with the highest reward. }
\begin{tabular}{cccccc} \toprule
    Task & PR2 & Jaco & Baxter & Sawyer & \textit{Success} \\ \midrule\midrule
    Itch Scratching & 80.9 & \textbf{443.2} & 83.3 & 131.2 & \textbf{68\%} \\
    Bed Bathing & 90.2 & \textbf{193.6} & 175.5 & 166.2 & \textbf{81\%} \\
    Feeding & \textbf{122.8} & 106.1 & 108.3 & 112.5 & \textbf{99\%} \\
    Drinking & \textbf{493.4} & 402.6 & 466.8 & 464.0 & \textbf{79\%} \\
    Dressing & -1.3 & 13.0 & 30.0 & \textbf{56.9} & \textbf{89\%}\\
	\bottomrule
\end{tabular}
\vspace{-0.4cm}
\end{table}

Overall, we observed that in almost all cases, a robot is able to provide better assistance and achieve a higher reward when learning to assist an active human who performs collaborative motions.
On average, the best performing robot achieved a 30.4\% increase in task success when compared to assisting a static human.
This is apparent when the robot assists with difficult to reach tasks, such as scratching underneath the upper arm, where an active human can rotate their arm to make it easier for the robot to reach and scratch the target location, as shown in Fig.~\ref{fig:scratching_coop}. Additional examples and results can be can be found in the supplementary video.

\section{Conclusion}

We presented Assistive Gym, an open source physics simulation framework for assistive robotics. Assistive Gym focuses on physical interaction between robots and humans. It models human physical capabilities and preferences for receiving assistance. We described baseline policies for four robots and six assistive tasks. We also demonstrated the use of Assistive Gym for benchmarking, for developing environments for multiple assistive tasks, and for comparing robots. Overall, we have shown that Assistive Gym is a promising open source framework for the development of autonomous robots that can provide versatile physical assistance.


\section*{Acknowledgment}

\small
\textit{We thank Alex Clegg, Yifeng Jiang, and Holden Schaffer for their guidance and assistance with this work. This work was supported by NSF award IIS-1514258 and AWS Cloud Credits for Research. Dr. Kemp owns equity in and works for Hello Robot, a company commercializing robotic assistance technologies.} 

\bibliographystyle{IEEEtran}
\bibliography{bibliography}

\balance

\end{document}